# Propagating construction-time knowledge quality into medical question answering: A framework grounded in clinical guidelines

Jie Hu[a], Junjie Wang[a], Shan Lu[b], Yifang Hu[c,d], Gong Cheng[e], Yun Liu[a,c,*]

[a] Department of Medical Informatics, School of Biomedical Engineering and Informatics, Nanjing Medical University, 101 Longmian Avenue, Jiangning District, Nanjing, 211166, China
[b] Department of Maternal and Child Care, The First Affiliated Hospital of Nanjing Medical University, 300 Guangzhou Road, Nanjing, 210029, China
[c] Department of Geriatrics, The First Affiliated Hospital of Nanjing Medical University, 300 Guangzhou Road, Nanjing, 210029, China
[d] Department of Clinical Medical Research, The Friendship Hospital of Ili Kazakh Autonomous Prefecture, 92 Stalin Street, Yining 835000, China
[e] State Key Laboratory for Novel Software Technology, Nanjing University, 163 Xianlin Avenue, Qixia District, Nanjing 210023, China

Corresponding author

Yun Liu. liuyun@njmu.edu.cn

## Abstract

Large language models have facilitated knowledge graph (KG) construction from clinical guidelines, but extracted triples vary in structural validity and evidential support. Meanwhile, graph-augmented question answering (QA) systems typically optimize query relevance during retrieval, with limited reuse of quality information produced during KG construction. This creates a disconnect between construction-time quality control and inference-time evidence use. We investigate whether construction-time triple quality can serve as a persistent signal for downstream evidence selection and presentation. We propose a quality-aware framework that models structural conformance (*SchemaConf*) and evidential support (*EvidScore*) as complementary dimensions and fuses them into a per-triple quality signal, $Q(t)$. Rather than using quality solely for filtering, the framework retains $Q(t)$ and derived quality tiers as graph attributes and propagates them into quality-weighted subgraph retrieval and tier-conditioned evidence prompting, while preserving passage-level provenance. Experiments on Chinese diabetes clinical guidelines show that the utility of the quality signal is distribution dependent. Under cross-version and cross-model shift, the fused $Q(t)$ provides stronger triple-quality discrimination than either component alone (AUC 0.748 vs. 0.703 for *EvidScore* and 0.645 for *SchemaConf*). In guideline-grounded QA, propagating construction-time quality reduces required-knowledge omission from 16.3% to 5.3% and conflicting outputs from 16.3% to 2.7%, with an evidence-grounded precision of 81.6% and near-zero invalid citations. Blinded clinician ratings favor the full framework over no retrieval (4.68 vs. 4.21 on a five-point scale) and approach the oracle condition (4.80), while cross-generator experiments show consistent trends.

## 1. Introduction

Clinical practice guidelines consolidate evidence-based diagnostic and therapeutic knowledge and provide an important foundation for clinical decision support, medical knowledge graph (KG) construction, and intelligent question answering (QA) systems (Nicholson & Greene, 2020; Jing et al., 2023). Organizing guideline knowledge into structured, machine-readable representations can facilitate evidence retrieval, knowledge reuse, and traceable medical QA. However, transforming free-text clinical guidelines into reliable KGs remains challenging because guideline documents are terminologically dense, semantically complex, heterogeneous in knowledge granularity, and often leave supporting-evidence boundaries implicit. These challenges motivate the present study, which uses Chinese diabetes clinical guidelines as the target domain for developing and evaluating the proposed framework.

Large language models (LLMs) have enabled scalable zero- and few-shot extraction of entities, relations, and structured facts with limited task-specific supervision (Wei et al., 2023; Wadhwa et al., 2023). LLM-driven pipelines have subsequently been applied to medical KG construction from multicenter clinical data and Chinese-language medical texts (H. Yang et al., 2025; L.-I. Wu et al., 2025). However, LLM-based extraction remains prone to errors in entity identification and granularity, type consistency, and relation extraction, including incorrect relation types and directions. The resulting triples therefore require refinement and alignment before they can be incorporated into a reliable KG. This creates a persistent quality-coverage trade-off: schema-based validation improves structural consistency and reduces structurally or semantically invalid triples, whereas less constrained extraction preserves broader coverage but requires additional validation (Hofer et al., 2024; Das et al., 2026). This trade-off motivates representing triple quality in graded form rather than relying solely on binary acceptance or rejection.

Graph-augmented QA provides a natural way to use such structured knowledge. Compared with conventional text-based retrieval-augmented generation (RAG) (Fan et al., 2024; Amugongo et al., 2025), graph retrieval-augmented generation (GraphRAG) leverages graph-structured knowledge through entities, relations, paths, and subgraphs, supporting relational retrieval and multi-hop reasoning (Edge et al., 2025; Peng et al., 2025). Representative methods include path-based and iterative graph exploration (Luo et al., 2024; Ma et al., 2025) and KG-guided or GNN-based retrieval (X. Zhu et al., 2025; Mavromatis & Karypis, 2025), while broader RAG research has explored adaptive or confidence-aware evidence selection (Asai et al., 2024; W. Wu et al., 2025). Medical variants further combine medical KGs, biomedical sources, and electronic health records (J. Wu et al., 2025; Zhao et al., 2025). Yet existing methods primarily estimate relevance, confidence, or source credibility during retrieval or generation, with limited attention to reusing quality information produced during KG construction. Consequently, construction-time triple reliability remains largely disconnected from downstream evidence selection, even though a highly relevant triple is not necessarily a reliable one.

To address the quality-coverage trade-off in KG construction and, more importantly, the disconnect between construction-time quality control and inference-time evidence use, we investigate a central question: can triple-level quality information produced during KG construction remain useful as a persistent signal for downstream evidence selection and use? We propose a quality-aware framework that preserves and propagates construction-time triple quality into medical question answering, as illustrated in Figure 1.

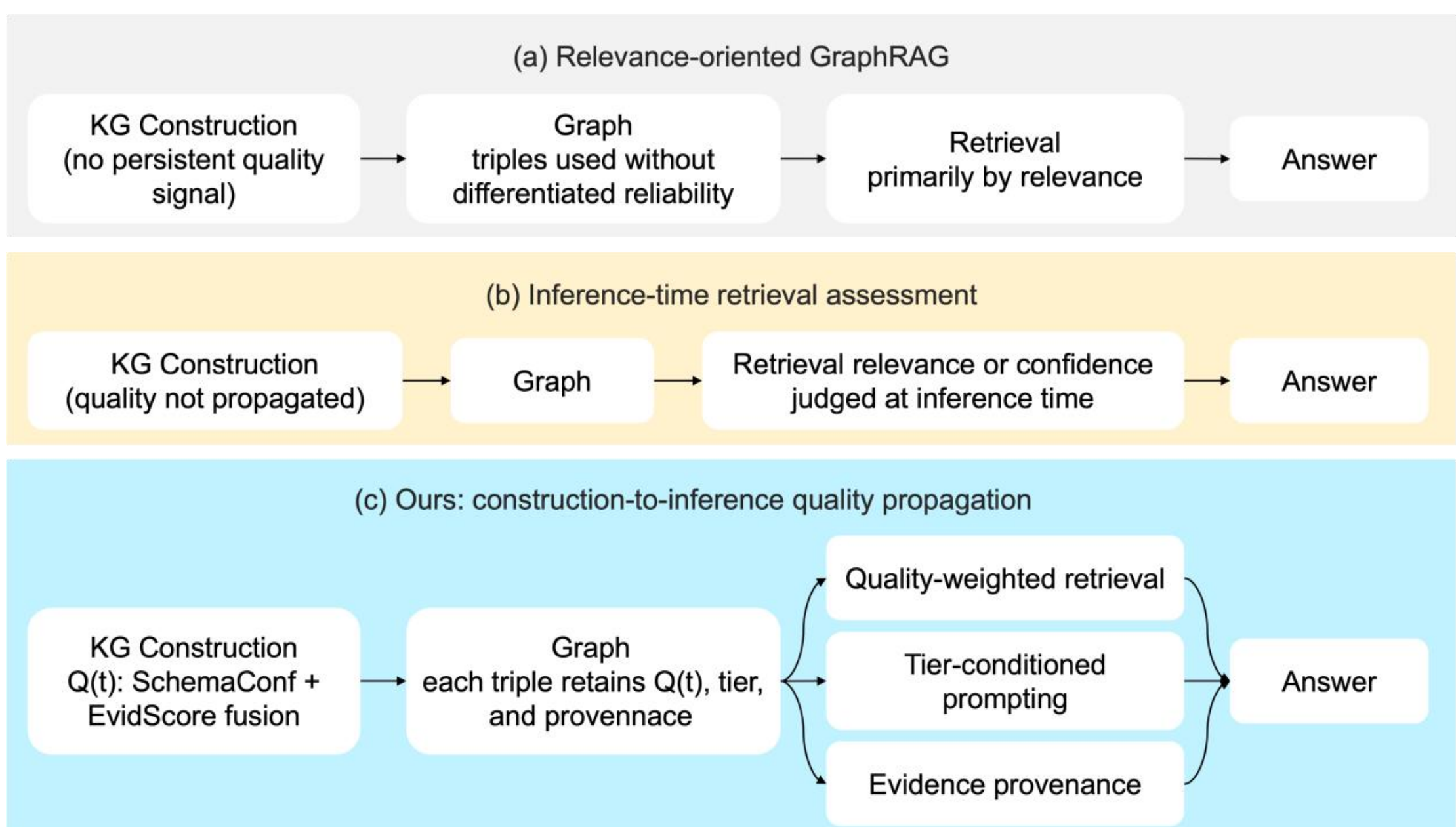

Fig. 1. Comparison of quality-signal use across retrieval-augmented QA settings. (a) Relevance-oriented GraphRAG uses graph triples without a persistent construction-time quality signal; (b) inference-time assessment methods estimate retrieval relevance or confidence during inference without propagating construction-time quality; and (c) our framework retains construction-time $Q(t)$ and tier information for quality-weighted retrieval and tier-conditioned prompting, while preserving passage-level evidence provenance.

During construction, the framework characterizes complementary aspects of triple quality through structural conformance (*SchemaConf*) and evidential support (*EvidScore*), which are fused into a continuous signal, $Q(t)$. Rather than using quality solely as a construction-time filtering criterion, $Q(t)$ is retained as a persistent triple attribute and used to organize knowledge into graded quality tiers. At inference, the same signal is carried forward into quality-weighted subgraph retrieval and tier-conditioned evidence prompting, while provenance metadata links graph evidence to the original guideline passages. The resulting framework therefore treats knowledge quality not as a local construction decision, but as information that can persist across KG construction, retrieval, and evidence use.

Importantly, we do not assume that propagating construction-time quality necessarily improves QA in all settings. Its downstream utility depends on whether the required knowledge is represented in the KG, successfully activated during retrieval, and meaningfully differentiated by the quality signal. We therefore evaluate not only whether quality propagation improves downstream performance, but also under what conditions the signal remains informative across construction and inference stages.

We instantiate the framework on Chinese diabetes clinical guidelines and evaluate it at both the construction and QA levels. The main contributions are:

- Graded triple-quality representation. We model structural conformance and evidential support as complementary dimensions of triple quality and retain their fused signal, $Q(t)$, beyond binary filtering. This enables uncertain but potentially useful knowledge to be organized into graded quality tiers while preserving reliability information for downstream use.
- Construction-to-inference quality propagation for graph-augmented QA. We operationalize construction-time quality through quality-weighted subgraph retrieval and tier-conditioned evidence prompting, with passage-level provenance retained for answer-to-source tracing. Through construction-side and downstream ablations, cross-version and cross-model evaluation, and blinded clinician assessment, we further examine when this propagation is effective and where its benefits are limited.

The remainder of the paper is organized as follows. Section 2 reviews related work on LLM-based clinical KG construction, knowledge-quality modeling, and KG-augmented medical QA. Section 3 presents the proposed quality-aware framework. Sections 4 and 5 describe the experimental setup and results, respectively. Section 6 discusses the framework’s applicability conditions and limitations, and Section 7 concludes the paper.

## 2. Related Work

### *2.1 LLM-based knowledge graph construction from clinical text*

Clinical KG construction traditionally relies on pipelines for named entity recognition, relation extraction, normalization, and triple assembly. Joint extraction methods integrated entity recognition and relation extraction, while unified frameworks extended structured prediction across multiple information extraction tasks (Zheng et al., 2017; Lu et al., 2022). Domain-pretrained language models further improved biomedical information extraction and related NLP tasks (Lee et al., 2020; Gu et al., 2021).

More recently, prompt-based LLMs have reduced reliance on task-specific training for information extraction. ChatIE formulated information extraction as a multi-turn zero-shot interaction with ChatGPT (Wei et al., 2023), whereas Wadhwa et al. (2023) evaluated LLMs for relation extraction under limited supervision. Beyond individual information extraction tasks, H. Yang et al. (2025) used GPT-4 to construct a sepsis KG from multicenter clinical data, guidelines, and public databases. L.-I. Wu et al. (2025) further investigated zero-shot construction of a medical KG from Chinese-language texts.

However, LLM-based construction still faces inconsistent entity granularity, weak schema or type compliance, relation errors, hallucination, and insufficient factual validation (L.-I. Wu et al., 2025; Maity & Saikia, 2025). Reviews of generative KG construction and information extraction similarly identify structural noncompliance and error propagation (Ye et al., 2022; Z. Zhang et al., 2025). Text2KGBench evaluates fact extraction accuracy, ontology conformance, and hallucination in ontology-guided KG generation (Mihindukulasooriya et al., 2023). These limitations are particularly consequential in medical settings, motivating quality-control mechanisms that combine schema-based validation, evidence assessment, and human review.

### *2.2 Knowledge quality assessment and confidence modeling*

Automatic KG construction increasingly incorporates quality control beyond extraction scale, including structural consistency, factual reliability, uncertainty management, and provenance tracking (Zhong et al., 2023; Hofer et al., 2024). Recent LLM-based studies further emphasize hallucination mitigation, external-source augmentation, and human validation during KG construction (Y. Zhu et al., 2024).

Existing work has examined KG refinement and error detection (Paulheim, 2017), alongside embedding-based confidence modeling for uncertain relational facts (X. Chen et al., 2019). More recent work has extended triple-confidence estimation to uncertainty-aware prediction (Yuqicheng Zhu et al., 2025).

For factual reliability, recent work has combined retrieved web evidence with multi-LLM verification and consensus voting for KG fact checking (Shami et al., 2025). However, model agreement alone does not establish factual correctness, motivating its combination with supporting evidence and, where appropriate, human validation.

Recent LLM-based frameworks further incorporate evidence verification and dynamic schema induction (Bao et al., 2026), as well as uncertainty assessment and multi-model validation (Das et al., 2026). More broadly, these quality estimates are typically used for construction-side filtering, error detection, or reporting, with limited attention to preserving and reusing them as per-triple signals in downstream retrieval and reasoning.

*2.3 Knowledge graph-augmented medical question answering*

RAG can mitigate knowledge staleness and reduce unsupported generation by supplying LLMs with retrieved external context (Fan et al., 2024). In healthcare, it has been applied to medical question answering, dialogue, and information retrieval tasks (Amugongo et al., 2025), while adaptive methods use self-reflection to determine when retrieval is needed and assess the relevance and supportiveness of retrieved evidence (Asai et al., 2024). Conventional text-based RAG generally represents documents as chunks and retrieves semantically relevant passages, but does not explicitly model the entities, relations, and multi-hop dependencies connecting them. GraphRAG instead organizes external knowledge as graphs and retrieves graph-structured evidence, such as entities, triples, paths, subgraphs, or graph summaries, to support generation and multi-hop reasoning (Edge et al., 2025; Peng et al., 2025; Kamalipour et al., 2026).

Representative graph-based RAG and KG-augmented reasoning approaches include relation-path planning and retrieval over KGs (Luo et al., 2024), iterative graph-text retrieval for deeper knowledge exploration (Ma et al., 2025), graph-indexed retrieval through graph propagation (Gutiérrez et al., 2024), KG-guided chunk expansion and context organization (X. Zhu et al., 2025), and graph-neural retrieval over KGs (Mavromatis & Karypis, 2025). Multi-source graph-based RAG further estimates graph- and node-level confidence to filter unreliable evidence and mitigate conflicts across knowledge sources (W. Wu et al., 2025).

In medicine, graph-augmented methods have been applied to evidence-grounded medical question answering (J. Wu et al., 2025), KG-guided diagnosis prediction from EHR narratives (Y. Gao et al., 2025), clinical reasoning over retrieved patient records (Zhao et al., 2025), and domain-specific applications such as Chinese diabetes QA (T. Yang et al., 2025).

Recent biomedical reviews further identify factual reliability, interpretability, and evidence grounding as important challenges for clinical deployment of LLM-KG systems, emphasizing grounded generation, factual validation, and human oversight in high-stakes medical applications (Xu et al., 2025; Murali et al., 2026; Agrawal et al., 2024; Gilbert et al., 2024).

Across much of this literature, retrieval relevance and final-answer performance remain central optimization targets. Although some systems estimate evidence confidence during retrieval or assess factual consistency after generation, construction-time knowledge quality—particularly structural validity and evidential support—has received limited attention as a persistent per-triple signal propagated across downstream retrieval, evidence organization, and prompting.

*2.4 From quality assessment to quality propagation*

Table 1 compares representative methods according to when quality or control signals are computed, their granularity, and how they are used downstream. Construction-oriented methods assess schema consistency, factual support, uncertainty, or model agreement, but primarily use these signals for validation, filtering, refinement, or quality assessment. Inference-oriented RAG and graph-based methods instead rely on query-dependent relevance, graph propagation, path scores, reflection tokens, or retrieval confidence. These two lines remain weakly connected: construction-derived reliability is rarely preserved as a persistent per-triple signal and reused to control downstream retrieval and evidence presentation. Our framework addresses this gap by computing structural and evidential quality during KG construction and propagating the resulting signal into downstream QA.

Table 1. Comparison of quality and evidence-control mechanisms across KG construction and downstream QA.

| Method | Signal or control mechanism | Stage of computation | Granularity | Primary use | Quality propagation |
|---|---|---|---|---|---|
| KG²RAG (X. Zhu et al., 2025) | Semantic similarity and KG-guided chunk expansion | Offline indexing + inference | Chunk/subgraph | Retrieval expansion and context organization | No |
| HippoRAG (Gutiérrez et al., 2024) | Personalized PageRank-based graph activation score | Inference | Node/passage | Graph-based retrieval ranking | No |
| GNN-RAG (Mavromatis & Karypis, 2025) | GNN-based graph relevance score | Inference | Node/subgraph | Graph-based retrieval | No |
| CRAG (Yan et al., 2024) | Retrieval evaluator confidence | Inference | Retrieved documents | Retrieval action | No |
| Self-RAG (Asai et al., 2024) | Reflection tokens | Inference | Passage/response | Retrieval and generation control | No |
| MultiRAG (W. Wu et al., 2025) | Graph- and node-level confidence | Inference | Graph/node | Retrieval and evidence filtering | No |
| Gao et al. (2025) | Composite triplet confidence and contextual metadata | Construction | Triple | Quality assessment and KG exploration | No |
| Das et al. (2026) | Uncertainty assessment and multi-model validation | Construction | Triple | Validation and refinement | No |

| FactCheck (Shami et al., 2025) | RAG-based LLM fact verification | Verification | Triple | Fact checking | No |
|---|---|---|---|---|---|
| Ours | *SchemaConf* + *EvidScore* -> Q(t) | Construction | Triple | Tiering, retrieval, prompting | Yes |

Note: Quality propagation indicates whether a quality signal computed during KG construction is retained and operationally reused in downstream retrieval or evidence presentation.

## 3. Proposed Framework

The proposed framework introduces a unified per-triple quality representation that is estimated during knowledge graph construction and reused in downstream question answering. Instead of treating KG construction and QA as independent stages, the framework establishes a continuous construction-to-inference pipeline in which the quality signal supports graded knowledge organization during construction and is subsequently reused for quality-weighted retrieval and tier-conditioned evidence prompting.

### *3.1 Overview and problem formulation*

**Overview.** As illustrated in Figure 2, the pipeline is organized into two phases connected by the per-triple quality signal $Q(t)$. The first phase, SEEK-GK (Schema-guided, Evidence-aware and Expert-in-the-loop Guideline Knowledge Graph Construction), transforms raw guideline text into a quality-annotated knowledge graph. Multiple LLMs extract candidate triples from guideline passages, and each triple is then evaluated along two complementary dimensions: structural validity through schema-conformance scoring (*SchemaConf*) and evidential reliability through evidence-aware scoring (*EvidScore*). The two scores are fused into the composite quality signal $Q(t)$, which organizes retained triples into reliability tiers $\tau(t) \in \{core, extended, exploratory\}$. The resulting tiered graph $G$ retains, for each retained triple, its quality score, tier label, and provenance links to the source guideline passages. The second phase, CDG-QA (Chinese Diabetes Guideline Question Answering), reuses $Q(t)$ to guide evidence selection and presentation at inference time. Given a question, entity linking identifies seed entities, from which a candidate subgraph is obtained through multi-hop expansion. Quality-weighted retrieval then ranks candidate triples by combining $Q(t)$ with query relevance, so that evidence reliability is considered alongside relevance rather than treating all retrieved triples equally. In parallel, dense passage retrieval supplies complementary guideline text. The combined evidence is organized by reliability tier and presented through a tier-conditioned prompt that prioritizes higher-tier evidence and encourages caution when support is limited. Supporting evidence for the generated answer can then be traced from retrieved triples to their source guideline passages through the stored *TextID* identifiers. The central design principle is therefore to preserve $Q(t)$ as an operational attribute of each retained triple and reused at inference time, without requiring joint training between the two phases.

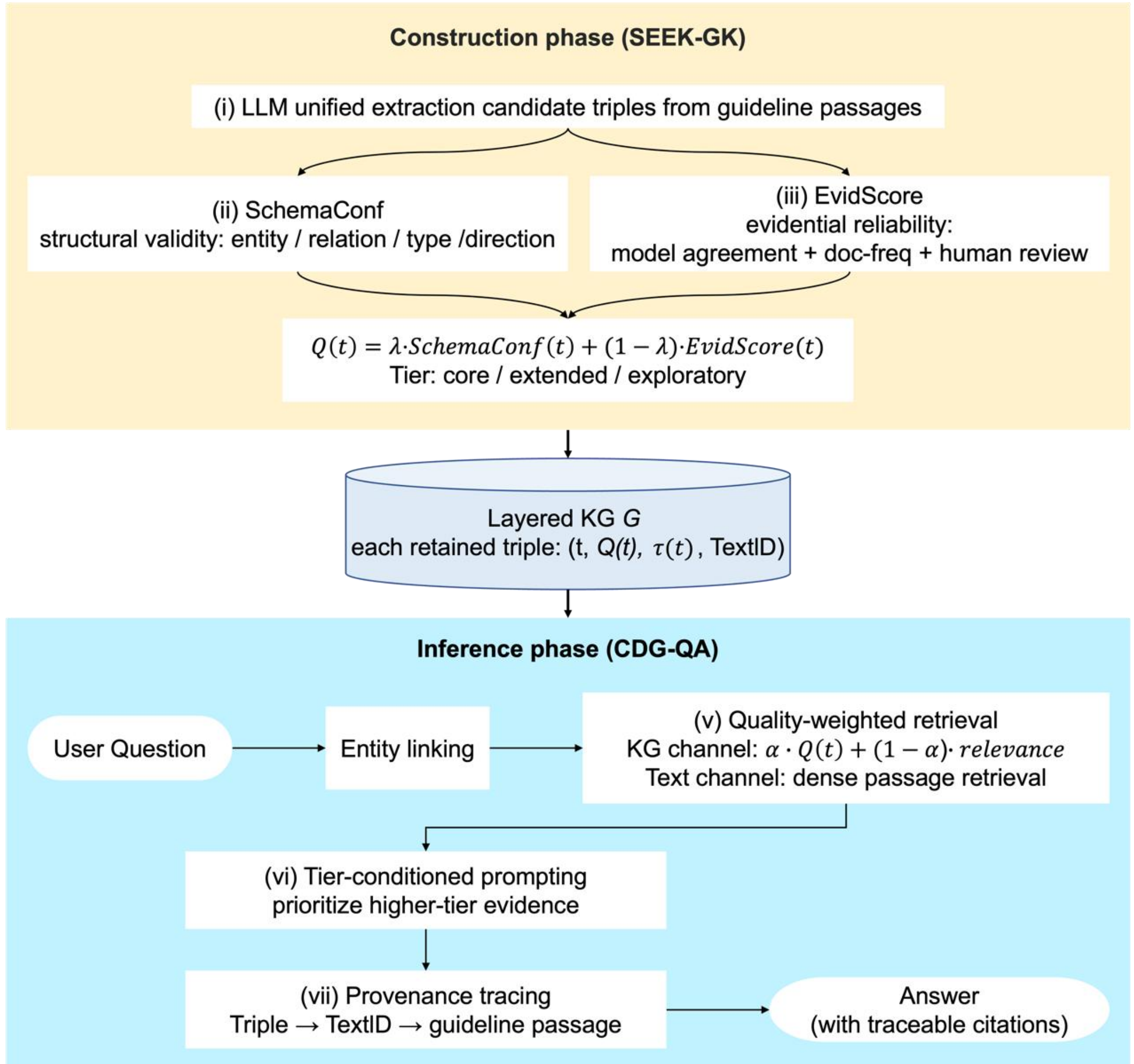

Figure 2. Overall construction-to-inference pipeline. The construction phase (SEEK-GK) estimates *SchemaConf* and *EvidScore*, fuses them into $Q(t)$, and organizes validated triples into quality tiers. The tiered graph $G$ retains $(t, Q(t), \tau(t), TextID)$ for each retained triple. The inference phase (CDG-QA) reuses $Q(t)$ in quality-weighted retrieval, tier-conditioned evidence prompting, and provenance tracing.

**Problem formulation**. Let the guideline corpus be $D=\{d_1, \ldots, d_m\}$, where each passage $d_k$ carries a stable identifier $TextID(d_k)$. LLM-based unified information extraction step produces a set of candidate triples $T=\{t=(h, r, o)\}$ where each triple $t$ is associated with a set of source passages $S_t \subseteq D$. For each candidate triple, we estimate two complementary quality dimensions:

- structural validity, represented by a schema-conformance score $SchemaConf(t) \in [0,1]$, measuring graded conformance to the domain schema with respect to entity, relation, type, and direction constraints (Section 3.3);
- evidential reliability, represented by an evidence-aware score $EvidScore(t) \in [0,1]$, aggregating multi-model agreement, passage-level

support, and human verification (Section 3.4).

The two dimensions are fused into a composite triple quality score:

$$Q(t) = \lambda \cdot SchemaConf(t) + (1-\lambda) \cdot EvidScore(t), \quad \lambda \in [0,1] \tag{1}$$

Each retained triple is then assigned a quality tier by thresholding $Q(t)$ (Section 3.5). Rather than applying a single hard threshold to all validated triples, the framework preserves graded quality differences among triples that pass the basic validity checks. Let $T^{+} \subseteq T$ denote the set of all triples retained after entity/relation normalization and quality scoring across all processed guideline editions. $T^{+}$ forms the runtime knowledge graph used at inference time; it includes both manually reviewed triples (from the primary construction corpus) and automatically scored triples (from incremental guideline editions processed by the same SEEK-GK quality framework). The resulting quality-annotated graph is represented as

$$G = \{ \left(t, Q(t), \tau(t), S_t\right) : t \in T^{+} \} \tag{2}$$

so that each graph triple is stored together with its quality score, tier, and provenance.

At inference time, given a question $q$, entity linking maps $q$ to a set of seed entities, and a candidate subgraph $T_q \subseteq T^{+}$ is obtained by multi-hop expansion. Rather than ranking triples solely by their association with the query entities, the framework combines the retained construction-time quality signal with a normalized query-triple relevance score $rel(t, q) \in [0,1]$:

$$score(t,q) = \alpha \cdot Q(t) + (1-\alpha) rel(t,q), \quad \alpha \in [0,1] \tag{3}$$

This formulation prioritizes evidence that is both relevant to the question and reliable according to the construction process (Section 3.7). The top-ranked triples, their quality tiers $\tau(t)$, and the corresponding guideline passages jointly form a tiered evidence context. A tier-conditioned prompt instructs the generator to prioritize higher-tier evidence and express appropriate caution when the available support is weak or uncertain. The supporting evidence remains traceable through the chain $t \rightarrow TextID \rightarrow$ guideline passage (Section 3.8).

**Design principle**. Existing approaches generally treat construction-time knowledge quality and inference-time evidence utility as separate concerns: construction-derived quality estimates are primarily used for validation, filtering, refinement, or quality assessment, whereas downstream QA typically relies on query-dependent relevance or confidence. The proposed framework instead preserves construction-time quality as a persistent per-triple attribute and reuses it during inference. Specifically, $Q(t)$ supports graded graph organization and quality-weighted retrieval, while its derived tier labels guide reliability-aware evidence prompting; provenance metadata is retained separately to support source tracing. In this way, construction-time knowledge quality remains operational throughout downstream evidence selection and use. The fusion weight $\lambda$ and retrieval weight $\alpha$ are determined empirically in the experiments (Sections 4-5).

*3.2 Guideline knowledge extraction and normalization*

Guideline documents are preprocessed using rule-based heuristics: non-knowledge content (headers, footers, references, and figure/table captions) is removed by pattern matching against document structure markers, and full-/half-width characters are normalized via Unicode conversion. Text is then segmented at paragraph, subsection, or semantic boundaries; fragments shorter than 50 characters are merged into the preceding segment, and fragments longer than 500 characters are split at sentence boundaries. Each fragment retains its heading-level path and a stable fragment identifier (*TextID*) used later for provenance tracking, evidence frequency counting, and error analysis. Three LLMs—GPT-4, GPT-4o, and GLM-4-Plus—are applied in parallel under a unified extraction prompt to generate candidate triples $T_{raw}$; duplicate triples within the same guideline are merged by exact-string matching on the text-normalized ($h$, $r$, $o$) tuple.

The canonical entity library $E$ is assembled from three complementary sources: (i) head and tail entities from the human-reviewed triple set $T_{\text{accepted}} \subseteq T_{\text{raw}}$, which is used for schema induction and is distinct from the correctness-labeled evaluation sets described in Section 4; (ii) manually curated NER outputs to recover entities not fully captured by relation extraction; and (iii) terminology entries from ICD-10/11, MedDRA, and CHPO that match guideline text. Synonym normalization proceeds through text cleaning, abbreviation expansion, format unification, and candidate synonym identification, with human confirmation of synonym equivalence and canonical names to form the library and synonym mapping $S_E$.

For relations, candidate synonym pairs are generated when two relations co-occur on the same ($h$, $o$) entity pair or have character-level similarity $\geq 0.5$, computed using the longest-common-subsequence (LCS) ratio. Each candidate pair ($r_1$, $r_2$) is then scored by:

$$SynonymScore(r_1, r_2) =$$

$$0.25 \cdot Cooccur_{norm} + 0.30 \cdot Jaccard_{HO} + 0.25 \cdot StrSim + 0.20 \cdot Substring \quad (4)$$

Here, $Cooccur_{norm}$ denotes normalized co-occurrence over shared ($h$, $o$) entity pairs; $Jaccard_{HO}$ is the Jaccard similarity between the argument-pair context sets $HO(r)=\{(\text{h},o) \mid (\text{h},r,o) \in T_{accepted}\}$; *StrSim* is the LCS-based character similarity; and *Substring* is a binary substring-containment indicator. The weights are empirically specified, with the slightly larger weight assigned to $Jaccard_{HO}$ to emphasize similarity in relation argument patterns. Candidate pairs are ranked and reviewed by domain experts. Relations differing only in surface modification (e.g., “may lead to” and “leads to” in Chinese: “可导致” and “导致”) are merged, whereas relations with directional or polarity differences are retained separately (e.g., indicated-for vs. contraindicated-for). Confirmed synonym clusters are aggregated using Union-Find to form the normalized relation set $R$ and mapping $S_R$. The schema layer then attaches

Domain/Range constraints derived from the empirical argument-type distributions of relations in $T_{accepted}$, followed by expert review and correction. Inverse-relation pairs $I_R$ are also recorded for direction normalization. Each retained triple preserves its source passage identifiers *TextID*($t$), enabling passage-level evidence assessment in Section 3.4 and answer-to-source provenance tracing in Section 3.8.

### *3.3 Schema-aware triple quality estimation (SchemaConf)*

*SchemaConf* measures how well a candidate triple conforms to the domain schema, extending binary schema compliance into a continuous, graded soft constraint so that structurally imperfect but potentially valid triples receive an intermediate score rather than being discarded. For a triple ($h$, $r$, $o$) it aggregates four sub-scores:

$$\mathrm{SchemaConf}(h, r, o) = w_e\, s_{entity} + w_r\, s_{relation} + w_t\, s_{type} + w_d\, s_{direction} \quad (5)$$

where $w_e + w_r + w_t + w_d = 1$. The weights are set at $w_e = 0.30$, $w_r = 0.25$, $w_t = 0.35$, and $w_d = 0.10$, assigning greater emphasis to entity and type compatibility; the weighting design is examined against uniform weighting and a type-dimension ablation in Section 5.2.

- $s_{entity} = (s_h + s_o)/2$, where $s_h$ and $s_o$ are the schema-matching scores of the head and tail entities. The entity-level scores $s_h$ and $s_o$, as well as the relation score $s_{relation}$, are determined using the same five-level matching scheme: (1) *exact match* (score 1.0): the string is found verbatim in the canonical library; (2) *synonym match* (score 0.95): the string maps to a canonical form via the human-confirmed synonym table; (3) *soft-suggest* (cosine similarity ≥ 0.85, score = similarity): embedding-based nearest-neighbor search indicates likely standardizability; (4) *soft match* (0.70 ≤ cosine similarity < 0.85, score = similarity): approximate match to a known concept; and (5) *no match* (cosine similarity < 0.70, score ≈ 0): treated as a new or out-of-schema concept. Embeddings are computed using a pre-built dense index (text-embedding-v3, 512-dimensional) over all canonical entities and relations; cosine similarity is used as the matching score.
- $s_{type}$ measures type compatibility against relation-specific Domain/Range constraints. Let *type*($h$) and *type*($o$) denote the entity-type sets for head and tail, respectively. Then:

$$s_{type} = \frac{1}{|C|}\sum_{c \in C} \delta_c, \quad \delta_c = \begin{cases} 1.0 & \text{if constraint } c \text{ is satisfied} \\ 0.5 & \text{if entity type is unknown (no penalty)} \\ 0.0 & \text{if type is known but violates } c \end{cases} \quad (6)$$

  Here, $C$ denotes the set of active Domain and Range constraints for relation $r$; if no such constraints are available, $s_{type} = 1.0$ by default so that missing schema information does not penalize the triple.
- $s_{direction}$ is a binary check for direction consistency with $I_R$: after applying canonical-direction normalization (swapping symmetric-relation arguments and resolving inverse mappings), $s_{direction} = 1.0$ for correctly-oriented triples and 0.0 for triples that cannot be resolved to a canonical direction.

The four dimensions target different failure modes (surface form, relation semantics, type composition, and direction), and their weighted sum yields a per-triple structural-validity score in [0, 1].

### *3.4 Evidence-aware triple quality estimation (EvidScore)*

*EvidScore* estimates evidential reliability from three complementary evidence signals:

$$EvidScore(t) = w_1\,E_{model}(t) + w_2\,E_{freq}(t) + w_3\,E_{human}(t) \qquad (7)$$

where $w_1+w_2+w_3=1$. $E_{model}$ reflects normalized multi-model agreement, measured as the proportion of models that independently extracted the triple; $E_{freq} = log(1+n)/\,log(1+N_{max})$ represents the log-normalized passage-level support frequency, where $n$ denotes the number of distinct *TextID*-linked guideline passages from which the triple is extracted and $N_{max}$ is the maximum support count observed in the corpus ($N_{max} = 6$ in our dataset); and $E_{human}$ encodes human verification. Among retained triples, manually accepted triples receive a full score of 1.0, whereas automatically validated but not manually reviewed triples receive a base value of 0.5, emphasizing human review without excluding automatically extracted knowledge. The default weights $w_1$=0.30, $w_2$=0.20, and $w_3$=0.50 are prespecified based on the relative reliability of the evidence sources and are not optimized on the evaluation data. They assign the highest contribution to human verification and a lower weight to support frequency to avoid disproportionately penalizing triples mentioned in relatively few passages.

One or more evidence dimensions may become inactive within a given batch—for example, when a batch lacks human verification labels, $E_{human}$ is identical for all triples; similarly, when per-triple passage counts are unavailable, $E_{req}$ is treated as inactive. Retaining fixed weights for inactive dimensions would compress the effective range of *EvidScore*. The framework therefore detects inactive (constant-valued) dimensions before batch scoring and redistributes their weights across the remaining active dimensions $A \subseteq \{1,2,3\}$ via proportional re-normalization:

$$w'_i = \frac{w_i}{\sum_{j \in A} w_j}, \quad i \in A; \qquad w'_i = 0, \quad i \notin A \qquad (8)$$

For example, when $E_{human}$ is inactive ($A = \{1,2\}$): $w'_1$=0.30/(0.30+0.20)=0.60, $w'_2$=0.20/(0.30+0.20)=0.40. The default weights are applied when all dimensions are informative. This adaptive re-weighting preserves the relative contribution of informative dimensions under varying evidence availability.

### *3.5 Quality fusion and tiered graph organization*

The two dimensions capture complementary aspects of triple quality—structural validity from schema conformance and evidential reliability from extraction agreement, passage-level support, and human verification — and are therefore modelled separately and fused into the composite quality score *Q*(*t*) defined in Eq. (1). We adopt a transparent weighted formulation so that the contribution of each quality dimension remains directly interpretable and adjustable without introducing additional learned parameters.

Triples are then organized into three quality tiers according to *Q*(*t*):

$$\tau(t) = \begin{cases} \text{core}, & Q(t) \geq \theta_{\text{core}}, \\ \text{extended}, & \theta_{\text{ext}} \leq Q(t) < \theta_{\text{core}}, \\ \text{exploratory}, & Q(t) < \theta_{\text{ext}}, \end{cases} \tag{9}$$

where $\theta_{core}=0.85$ and $\theta_{ext}=0.70$ are selected on the in-distribution calibration set based on the precision-coverage trade-off described in Section 5.2. The resulting tiers enable downstream tasks to select evidence according to their reliability requirements. Crucially, low-scoring triples are tiered rather than removed, preserving potentially useful boundary knowledge while indicating lower confidence.

The organization also supports lightweight incremental maintenance. When an exploratory or extended triple is later confirmed correct through human review, its $E_{human}$ increases from the base value to the full score, thereby improving *EvidScore* and *Q*(*t*) and potentially promoting the triple to a higher tier. A triple confirmed incorrect during review is removed. Because quality estimation is performed at the triple level, additional verification or localized guideline revisions require recomputing quality scores only for affected triples rather than rebuilding the entire graph.

### *3.6 Propagating construction-time quality to downstream question answering*

Sections 3.3-3.5 assign each retained triple a quality score *Q*(*t*) and a tier label τ(t), which are stored as persistent attributes of the tiered graph *G*. The central design decision is to retain these attributes with the graph and reuse them at inference time rather than limiting them to construction-side quality control. This provides the mechanism by which construction-time knowledge quality is propagated into downstream question answering.

Quality propagation operates through two inference-time interfaces:

- Retrieval interface. The quality score *Q*(*t*) enters the candidate-ranking function (Eq. 3), so that evidence reliability is considered alongside query relevance during subgraph selection (Section 3.7).
- Prompting interface. The tier label τ(t) conditions how triples are presented to the generator — core-tier triples as primary evidence and exploratory-tier triples as lower-confidence evidence requiring cautious use — so that evidence reliability influences how retrieved knowledge is used during answer generation (Section 3.8).

In parallel, provenance links retained during construction associate each triple with its source *TextID* values, allowing retrieved graph evidence to be traced back to the corresponding guideline passages (Section 3.8).

Crucially, this propagation requires no joint training or shared parameters between construction and QA. $Q(t)$ and $\tau(t)$ carry construction-time quality information into inference, while *TextID* preserves source provenance. The downstream pipeline consumes these attributes without modifying the underlying generator. Sections 3.7 and 3.8 detail their inference-time use.

### *3.7 Quality-aware graph and text retrieval*

At inference time, a question $q$ is linked to seed entities through a three-stage cascade: (1) exact substring matching—KG entities found verbatim in the question are linked with confidence 1.0; (2) segmentation matching—the question is tokenized using jieba with all KG entity names pre-registered as custom words to improve boundary accuracy, and tokens matching the entity library are linked with confidence 0.95; and (3) fuzzy matching—applied only when fewer than three candidates are found in the earlier stages, using a SequenceMatcher ratio ≥ 0.75 between candidate question spans and KG entity names. Matched entities serve as seed nodes, from which a candidate subgraph $T_q \subseteq T^+$ is obtained through up to two-hop expansion over $G$.

Instead of ranking triples solely by their association with the query entities, the framework combines the retained construction-time quality signal with an entity-link-based relevance score (Eq. 3). Specifically, $rel(t, q)$ is defined as the maximum entity-linking confidence among entities in $t$ matched to the question, so that triples containing more confidently linked query entities receive higher relevance scores. The top-ranked triples are returned as KG evidence.

In parallel, guideline passages are encoded using BGE-M3 (1024-dimensional, L2-normalized) and ranked by cosine similarity to the question, with the top-5 passages returned as complementary textual evidence supplying conditions, qualifiers, and explanatory context that structured triples may omit. The two channels form a dual-source evidence set in which each KG triple retains its quality score $Q(t)$, tier label $\tau(t)$, and source *TextID*($s$).

### *3.8 Tiered evidence prompting and provenance tracing*

The retrieved evidence is presented to the generator through a tier-conditioned prompt. Triples are grouped by tier with explicit quality annotations and usage rules: core-tier knowledge is prioritized as the primary evidence basis for answer generation; extended-tier knowledge provides supplementary support; and exploratory-tier knowledge is explicitly marked as requiring verification and should not be presented as a definitive conclusion. When neither the KG channel nor the passage channel provides sufficient support, the prompt instructs the model to indicate the evidence limitation rather than speculate. For high-risk relations such as contraindications, dosages, and adverse reactions, exploratory-tier evidence is restricted to a reference-only role and core-tier knowledge is preferred, so that reliability information regulates the use of uncertain evidence in high-risk scenarios. Finally, supporting graph evidence used in answer generation can be traced through $t$ → TextID → guideline passage, enabling users to verify retrieved evidence against the source guideline. Figure 3 summarizes the resulting quality-aware dual-channel retrieval and tier-conditioned prompting.

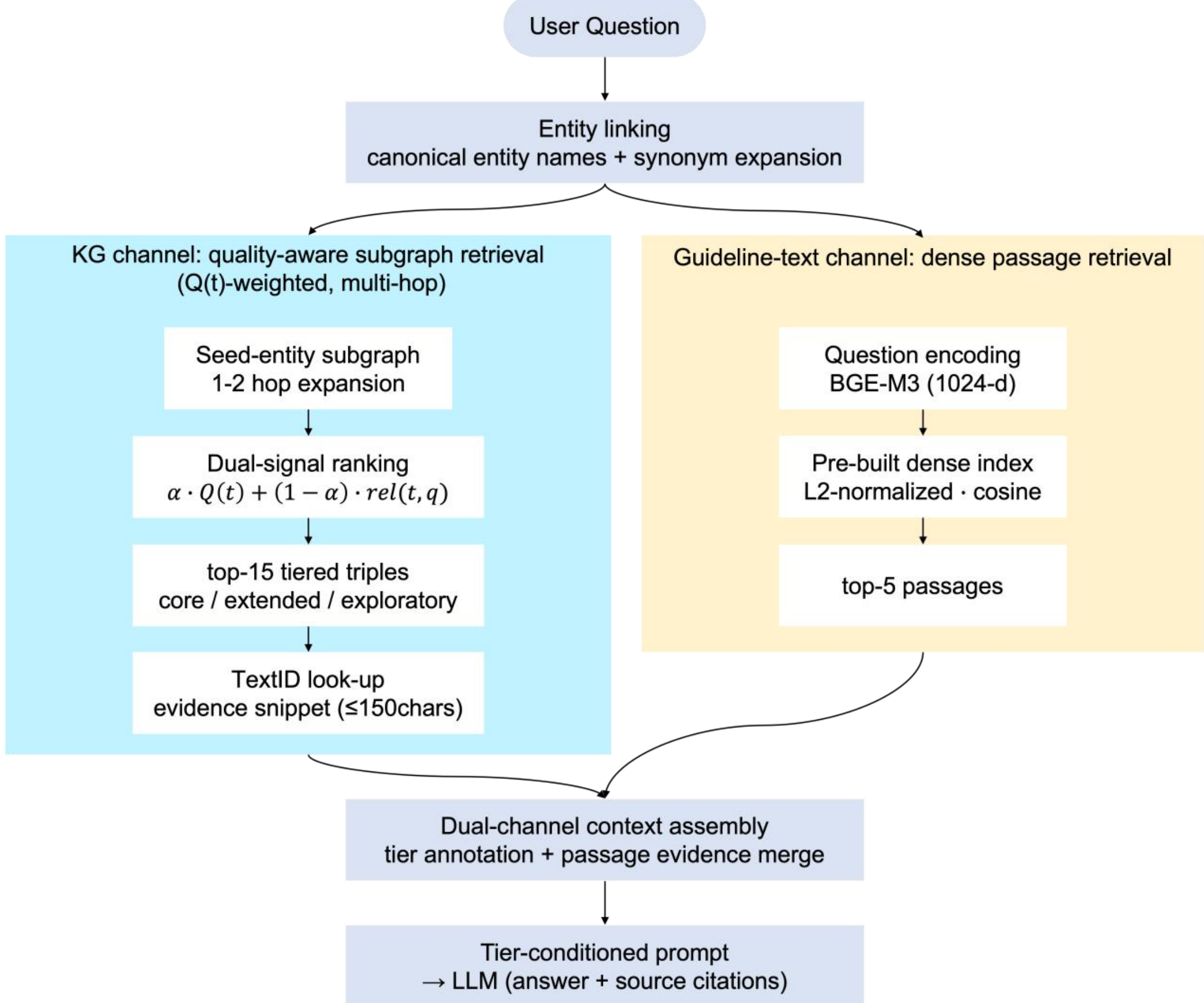

Figure 3. Quality-aware dual-channel retrieval and tier-conditioned evidence prompting. The KG channel performs $Q(t)$-weighted multi-hop subgraph retrieval while retaining tier and provenance information; the guideline-text channel supplies complementary context through dense passage retrieval. The two evidence channels are then merged and presented to the LLM through a tier-conditioned prompt.

## 4. Experimental Setup

We evaluate the framework at two levels: (i) knowledge graph construction quality, including the validity and reliability of extracted triples, and (ii) guideline-based medical question answering performance. The construction and inference evaluations share the same entity library, relation set, and schema constraints, while the quality scores and tier labels produced during construction are retained for downstream QA.

*4.1 Guideline corpus and knowledge graph construction data*

The primary corpus is compiled from a collection of authoritative Chinese diabetes clinical guidelines (Supplementary Table S1). Diabetes mellitus is a high-burden chronic disease with growing global prevalence and substantial demand for evidence-based decision support (Genitsaridi et al., 2026), making it a suitable domain for evaluating the proposed framework. After preprocessing, the text is segmented into 1,905 passages (mean length ≈ 300 characters), each retaining its heading path and fragment identifier. Three LLMs — GPT-4, GPT-4o, and GLM-4-Plus — perform unified information extraction in parallel under a shared prompt, producing 95,538 raw triples that reduce to 94,987 after intra-guideline deduplication. Relation normalization and subsequent Domain/Range specification result in 1,076 type-constraint rules over 538 relations. Following iterative expert review and schema-guided normalization, 15,636 unique triples are retained as the primary construction resource. Incorporating additional triples from a 2024 guideline edition (Chinese Diabetes Society, 2025) scored by the same SEEK-GK quality framework yields the final tiered graph of 34,672 triples used in the QA stage. The three extraction models show limited overlap (Figure 4), indicating substantial diversity across model outputs and motivating multi-model fusion to broaden candidate coverage.

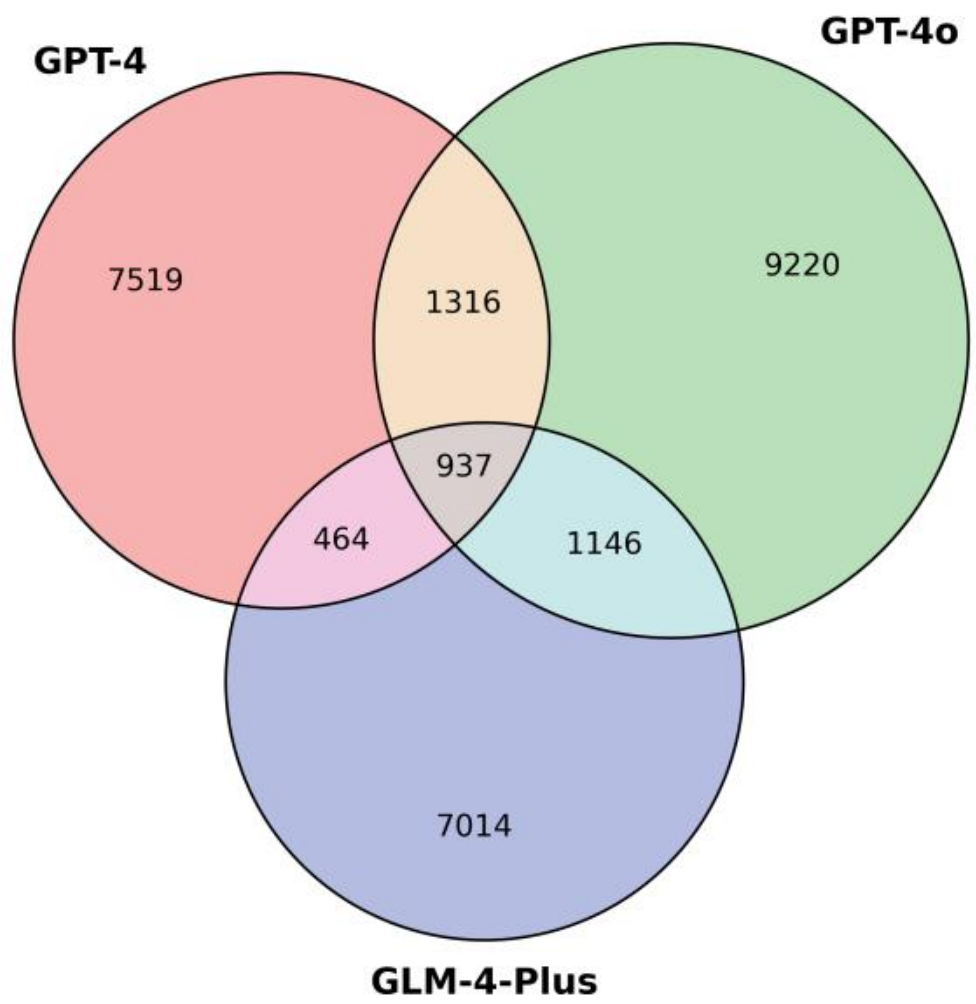

Figure 4. Accepted triple overlap across the three extraction models applied to the primary construction corpus. The limited pairwise overlap and small three-model intersection (937 triples) illustrate the diversity of accepted triples contributed by the extraction models.

To evaluate robustness under distribution shift, we additionally build a cross-version test set from the 2024 edition of the guideline. The text is segmented into 370 passages, and triples are extracted using three models — Qwen3-Max, GLM-4-Plus, and DeepSeek-V3 — two of which are not used in the primary construction corpus. The cross-version test set thus differs from the construction corpus in both guideline version and extraction-model composition, approximating the realistic setting of continued graph maintenance. A stratified sample of 1,888 triples from this set is manually labeled for factual correctness by domain experts, providing gold correctness labels for quality discrimination evaluation (Section 5.2). For the in-distribution condition, a stratified sample of 300 triples drawn from the primary construction corpus is separately labeled for factual correctness by domain experts, forming an in-distribution calibration set used both for threshold selection (Section 3.5) and for characterizing tier-precision stability under varying λ (Section 5.2). We refer to these two labeled sets as the calibration set (in-distribution, n=300) and the cross-version test set (distribution-shift, n=1,888) throughout.

### *4.2 Medical question answering dataset*

Because the proposed evaluation requires questions to be associated with verifiable KG evidence paths and passage-level TextID identifiers, we construct a guideline-based QA dataset tailored to this setting. Questions are constructed from predefined evidence paths in the retained KG and subsequently refined through expert curation. The resulting dataset contains 440 questions across seven types: single-hop forward, single-hop inverse, multi-hop, entity comparison, constraint, guideline-overview, and long-tail questions. Each question is associated with its source TextID(s) and corresponding guideline passages, and all questions are reviewed by clinicians for wording clarity and answer consistency. Existing benchmarks do not jointly provide the construction-time triple-quality information, guideline-level provenance, and predefined evidence grounding required by this evaluation protocol.

Of the 440 questions, 301 have explicitly predefined required KG paths suitable for path-anchored evaluation, comprising single-hop forward, single-hop inverse, multi-hop, and long-tail questions. All required paths for these questions are present in the runtime graph, ensuring that path-level evaluation measures retrieval and evidence-use performance rather than graph coverage. The remaining 139 questions cover entity-comparison, constraint, and guideline-overview scenarios and are included in the broader QA evaluation.

### *4.3 Baselines and implementation*

Construction baselines. At the construction level, we evaluate two aspects. First, the proposed tiered retention strategy is compared with schema-based hard-filtering baselines of increasing strictness. Second, the discriminative ability of the fused score $Q(t)$ is compared with its individual components, *SchemaConf* and *EvidScore*. For clarity, we distinguish two three-way partitions used in this study: the *SchemaConf* class (validated/candidate/uncertain; Section 5.1), a single-dimension partition computed from *SchemaConf* for construction-side analysis, and the fused-score quality tier $\tau(t)$ (core/extended/exploratory, Section 3.5), which governs the retained graph and downstream retrieval. The two partitions are related but not identical.

QA baselines / ablations. Using the same question set and generation model, we evaluate a nested set of retrieval and prompting configurations: ‘no_rag’ (LLM only), ‘text_only’ (passage RAG), ‘kg_only’ (graph only), ‘no_qt’ (dual-channel retrieval without $Q(t)$-based ranking), ‘qt_no_tier’ (quality-weighted retrieval without tier-conditioned prompting), ‘full’ (quality-weighted retrieval with tier-conditioned prompting), and ‘oracle_kg’ (predefined required KG evidence directly injected as an upper-bound condition). This design isolates the contribution of (a) propagating construction-time quality to retrieval ranking (‘no_qt’→‘qt_no_tier’) and (b) tier-conditioned evidence prompting (‘qt_no_tier’→‘full’).

Implementation. Passage retrieval is performed over the full 2,275-passage guideline library (the primary construction corpus combined with passages from the newer guideline edition), using BGE-M3 dense embeddings (1024-dimensional, L2-normalized) with cosine similarity and returning the top-5 passages. KG retrieval performs two-hop subgraph expansion and returns the top-15 triples ranked by the quality-aware retrieval score defined in Eq. (3), with fusion weight $\lambda$=0.3 (Eq. 1) and retrieval weight $\alpha$=0.4 (Eq. 3). The latter is prespecified so that entity-link-based relevance remains the dominant component while $Q(t)$ provides a quality-aware re-ranking signal; sensitivity to $\lambda$ is examined in Section 5.2. Answer generation uses GLM-4-Plus with deterministic decoding (temperature = 0.0).

To assess whether the observed improvements generalize across different generators, we additionally evaluate a cross-LLM subset of 60 KG-anchored questions, stratified across the four KG-anchored question types (15 each for single-hop forward, single-hop inverse, multi-hop, and long-tail; random seed 42), using two alternative generators, Qwen-Max and DeepSeek-V3. Retrieval, prompting, and evaluation protocols are kept unchanged from the main evaluation so that only the generator varies. The evaluation judge is GLM-4-Plus, identical to the main evaluation. All retrieval indices are precomputed and cached.

### *4.4 Evaluation metrics*

Construction. We report the number of correctly retained triples (knowledge preservation) and tier precision, estimated by inverse-probability-weighted sampling of manually reviewed triples. ROC-AUC is used to characterize the discriminative power of quality scores, with variation across fusion weights used to assess score stability on the in-distribution calibration set and the cross-version test set.

QA. Following an evidence-grounded, KG-anchored protocol, we report: Required-Knowledge Omission Rate (KOR) — the proportion of KG-anchored questions classified as MISSING with respect to their pre-specified required evidence; Conflict Rate (CR) — the proportion of questions whose answers contradict the standard KG or guideline evidence; Evidence-Grounded Precision (EGP)—the degree to which valid citations support their corresponding claims, with partial support receiving half credit; the auxiliary Invalid-Citation Rate (ICR) — the proportion of cited identifiers that cannot be resolved; and Unsupported-Claim Rate (UCR) — a weighted proportion of factual claims lacking sufficient evidential support, with partially supported claims assigned half weight. Metric denominators are anchored to predefined evidence or answer-claim populations rather than to selectively retrieved system outputs. The metrics are formally defined as follows:

$$\mathrm{KOR} = \frac{|\{\mathrm{q} \in \mathrm{Q_{KG}} | \mathrm{cov(q)} = \mathrm{MISSING}\}|}{|\mathrm{Q_{KG}}|} \tag{10}$$

$$\mathrm{CR} = \frac{N_{CONFLICT}}{N_{CONFLICT} + N_{NO_CONFLICT}} \tag{11}$$

$$\mathrm{EGP} = \frac{\mathrm{N_{SUPPORT}} + 0.5\,\mathrm{N_{PARTIAL}}}{\mathrm{N_{SUPPORT}} + \mathrm{N_{PARTIAL}} + \mathrm{N_{MISMATCH}}} \tag{12}$$

$$\mathrm{ICR} = \frac{\mathrm{N_{NOT_FOUND}}}{\mathrm{N_{allcitations}}} \tag{13}$$

$$\mathrm{UCR} = \frac{\sum_{\mathrm{q}} (\,0.5\,\mathrm{n_{par,q}} + \mathrm{n_{unsup,q}})}{\sum_{\mathrm{q}} (\,\mathrm{n_{sup,q}} + \mathrm{n_{par,q}} + \mathrm{n_{unsup,q}})} \tag{14}$$

where $Q_{KG}$ denotes the KG-anchored question set (n=301) and $cov(q) \in \{COVERED, PARTIAL, MISSING\}$ denotes question-level coverage of the pre-specified required triples for question *q*. A question is counted as an omission only when its coverage label is *MISSING*; *PARTIAL* (incomplete but non-zero coverage) does not constitute omission. EGP excludes *NOT_FOUND* identifiers, which are reported separately through ICR; *PARTIAL* receives half credit to represent incomplete but non-zero evidential support. UCR is computed over all verifiable factual claims in the answers, classified per-claim as supported (*sup*), partially supported (*par*), or unsupported (*unsup*) by the provided evidence.

Required-knowledge inclusion for KOR is assessed using a two-stage protocol: keyword matching is performed first, followed by GLM-4-Plus semantic review for borderline cases (output: *COVERED* / *PARTIAL* / *MISSING*). Semantic judgments for conflict, evidence support, and unsupported claims are assessed by GLM-4-Plus using fixed prompts restricted to predefined labels with deterministic decoding (temperature = 0). For evidence-support judgments, unparseable outputs default to *MISMATCH*; for conflict and unsupported-claim judgments, unparseable outputs default to *UNCLEAR* and are excluded from the corresponding metric denominators. LLM-based judgments are treated as automatic evaluation signals rather than substitutes for clinical assessment; Section 4.5 therefore provides an independent blinded clinician evaluation of answer quality. Deterministic sub-metrics that do not require semantic judgment (e.g., citation-ID validity) are computed by scripts. Paired significance for question-level binary outcomes (KOR, CR) is assessed using McNemar's exact test on questions with valid labels in both compared conditions, with Holm-Bonferroni correction within metric families; 95% continuity-corrected Wilson (Wilson-CC) confidence intervals are reported for the main mode comparisons in Section 5.3.

*4.5 Clinician evaluation protocol*

To independently validate answer quality beyond the automatic evaluation pipeline, two clinicians specializing in diabetes care conduct blinded evaluations of 100 stratified questions across three system outputs (‘no_rag’, ‘full’, and ‘oracle_kg’). Each answer is rated on four 5-point Likert-scale dimensions — accuracy, completeness, hallucination control, and overall quality—yielding 600 clinician-answer rating instances in total (100 questions × 3 systems × 2 raters). System identities are hidden during rating. We report Wilcoxon signed-rank tests for paired system comparisons and characterize inter-rater consistency using Pearson correlation and the proportion of ratings differing by no more than one point. This evaluation serves as an independent clinical cross-check of the automatic indicators rather than the sole basis for the conclusions.

## 5. Results

We organize the results around three questions: (i) whether the proposed tiered quality modeling preserves correct knowledge relative to hard filtering and provides reliable quality discrimination (Sections 5.1-5.2); (ii) whether propagating construction-time quality improves guideline-based question answering and which components contribute to the gains (Sections 5.3-5.4); and (iii) whether the resulting answers receive favorable clinician assessments, remain evidence-traceable, and generalize across generators (Sections 5.5-5.6). We then illustrate the end-to-end mechanism through a case study and provenance analysis (Section 5.7), examine which error types the quality-scoring mechanism can and cannot distinguish on the manually labeled cross-version test set (Section 5.8), and synthesize the conditions under which quality propagation is most effective (Section 5.9).

*5.1 Schema-aware knowledge preservation versus hard filtering*

Compared with binary schema filtering, graded *SchemaConf*-based retention preserves substantially more correct knowledge. On the cross-version test set, the strictest schema-based hard-filtering baseline retains 6,378 triples (≈3,155 estimated correct), whereas SEEK-GK retains 13,244 triples in the *SchemaConf*-validated class (≈6,304 estimated correct)—approximately twice as many correctly retained triples. The *SchemaConf* candidate class contains additional correct knowledge, whereas the uncertain class exhibits near-zero precision (< 1%), indicating that *SchemaConf* separates structurally weak candidates while preserving useful knowledge near the validation boundary. When the cross-version test set is stratified by model support, the largest preservation gains over hard filtering occur in the low-support band, where strict filtering removes the greatest amount of otherwise retainable knowledge.

*5.2 Quality-score discrimination under distribution shift*

The discriminative behavior of $Q(t)$ differs between in-distribution and distribution-shift settings. On the in-distribution calibration set (n = 300), varying the fusion weight λ mainly shifts the decision thresholds with little change in triple ranking: the precision of the $Q(t)$-based tiers remains stable (core ≈ 93.9-94.0%, extended ≈ 78.0-78.4%, with a stable 15.5-16.0-point gap), and the AUC is nearly constant across λ. In this setting, the resulting $Q(t)$ ranking remains close to that produced by *EvidScore* alone, leaving limited room for *SchemaConf* to provide additional discrimination. On the cross-version, cross-model test set, however, the fused $Q(t)$ achieves higher discrimination (AUC 0.748) than either *EvidScore* (0.703) or *SchemaConf* (0.645) alone; DeLong's test confirms that both gaps are statistically significant ($Q(t)$ vs. *EvidScore*: $\Delta$ AUC = 0.045, $z$ = 9.31, $p < 0.001$; $Q(t)$ vs. *SchemaConf*: $\Delta$AUC = 0.103, $z$ = 8.36, $p < 0.001$; n = 1,888). We therefore interpret the schema component not as a universal improvement, but as a complementary structural constraint whose added value emerges under distribution shift, a setting that approximates continued KG maintenance across guideline updates and extraction-model changes. The two quality dimensions show weak correlation on the labeled cross-version test set (Pearson $r$ = 0.16; Figure 5), suggesting complementary rather than redundant signals; the discrimination gain under shift is shown in Figure 6.

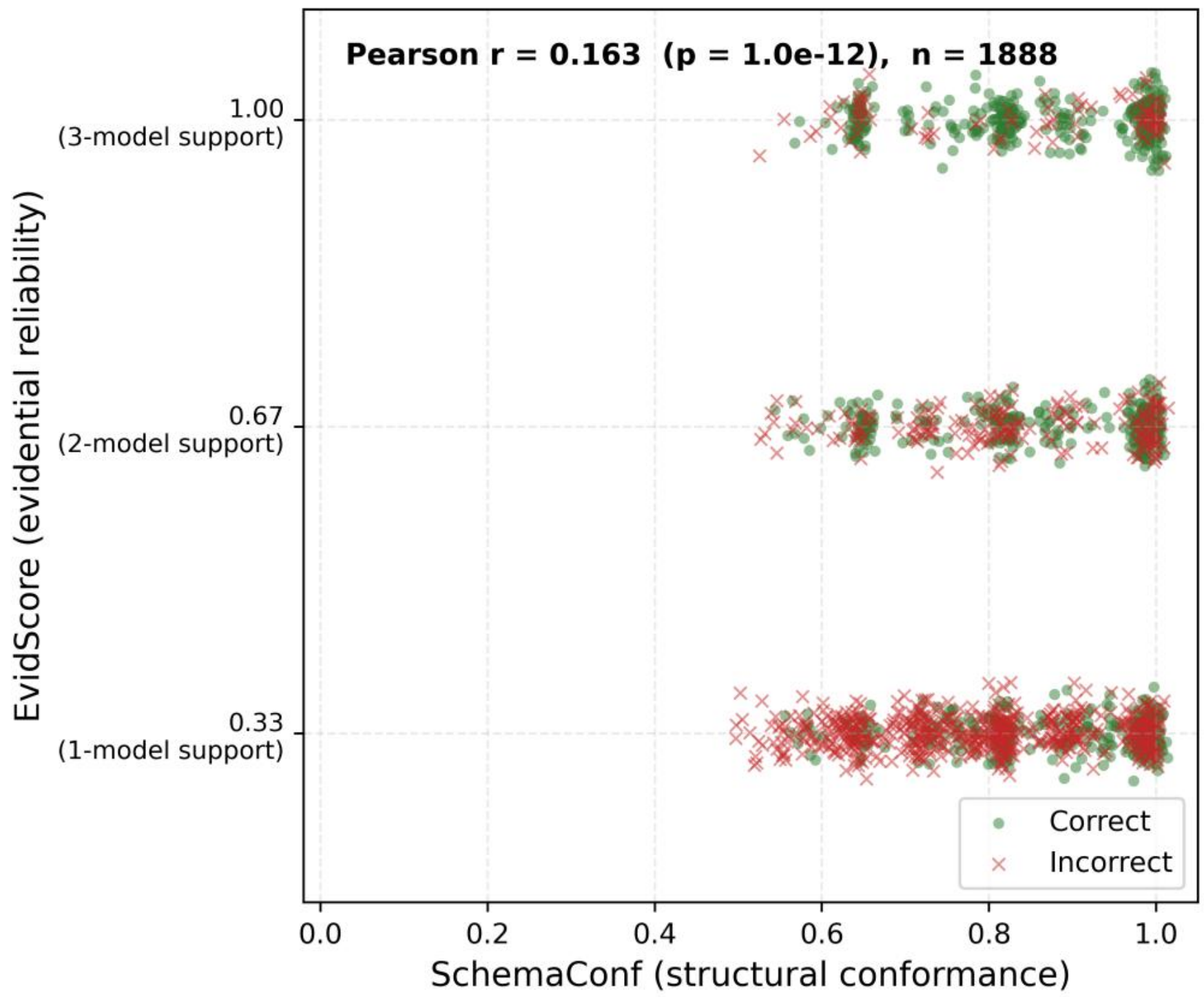

Figure 5. *SchemaConf* and *EvidScore* show weak correlation (Pearson $r$ = 0.16 on the cross-version test set), suggesting that structural validity and evidential reliability capture complementary quality signals and motivating their separate modeling.

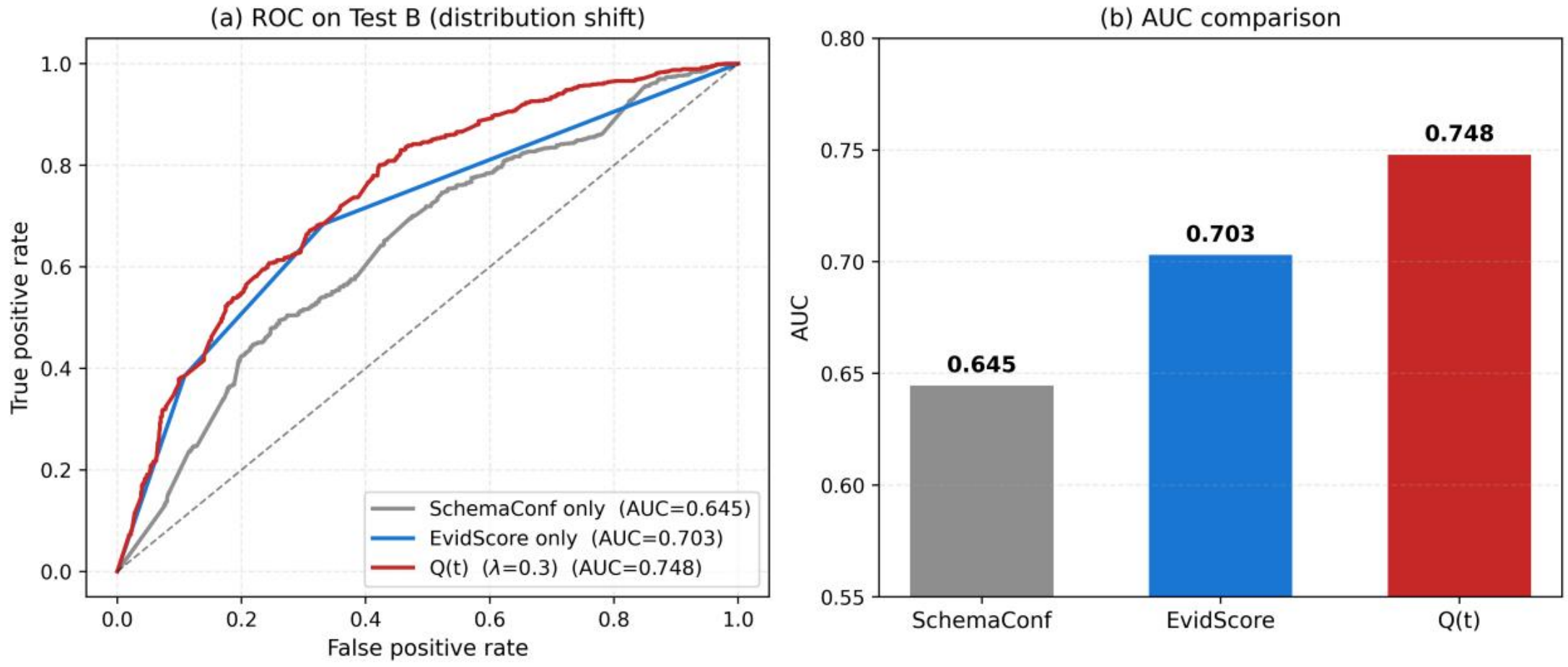

Figure 6. Discrimination under distribution shift (cross-version test set). The fused $Q(t)$ (AUC 0.748) outperforms *EvidScore* (0.703) and *SchemaConf* (0.645) alone; the schema component contributes complementary discriminative power under cross-version, cross-model shift.

*SchemaConf* weight ablation. To examine the proposed weighting ($w_e$ = 0.30, $w_r$ = 0.25, $w_t$ = 0.35, $w_d$ = 0.10), we compare against two alternatives on the cross-version test set (n = 1,888): (i) Uniform weighting ($w$ = 0.25 for all four dimensions) and (ii) No-$s_{type}$ (replacing the type-compatibility score with a neutral constant 0.5). Under uniform weighting, the validated-class size increases from 13,244 to 15,619 while IPW precision drops from 47.6% to 43.4%. This is consistent with the direction dimension contributing little variation after inverse-relation normalization, while uniform weighting reduces the relative contribution of type compatibility. Removing $s_{type}$ yields a similar effect: precision falls to 44.7% while the uncertain class vanishes entirely, highlighting the important role of type compatibility in structural discrimination. Overall, the proposed weighting maintains higher validated-class precision than either alternative while preserving a meaningful separation between confidence classes. We note that these weights were not optimized on the evaluation data; the ablation is a post-hoc robustness check rather than a tuning procedure.

### *5.3 Overall guideline-based QA performance*

The full framework substantially improves answer evidence quality relative to the no-retrieval baseline. Required-knowledge omission decreases from 16.3% to 5.3% (95% Wilson-CC CI: 3.2-8.6%; McNemar $p < 0.001$), and the conflict rate decreases from 16.3% to 2.7% (95% Wilson-CC CI: 1.3-5.4%; $p < 0.001$) (Figure 7a,b). Evidence-grounded precision reaches 81.6% with a near-zero invalid-citation rate, while the unsupported-claim rate decreases from 72.1% in ‘no_rag’ to 28.8% in the ‘full’ framework (Figure 7c,d). For supported statements, evidence can be traced through triple → TextID → guideline passage, providing an explicit answer-to-source path.

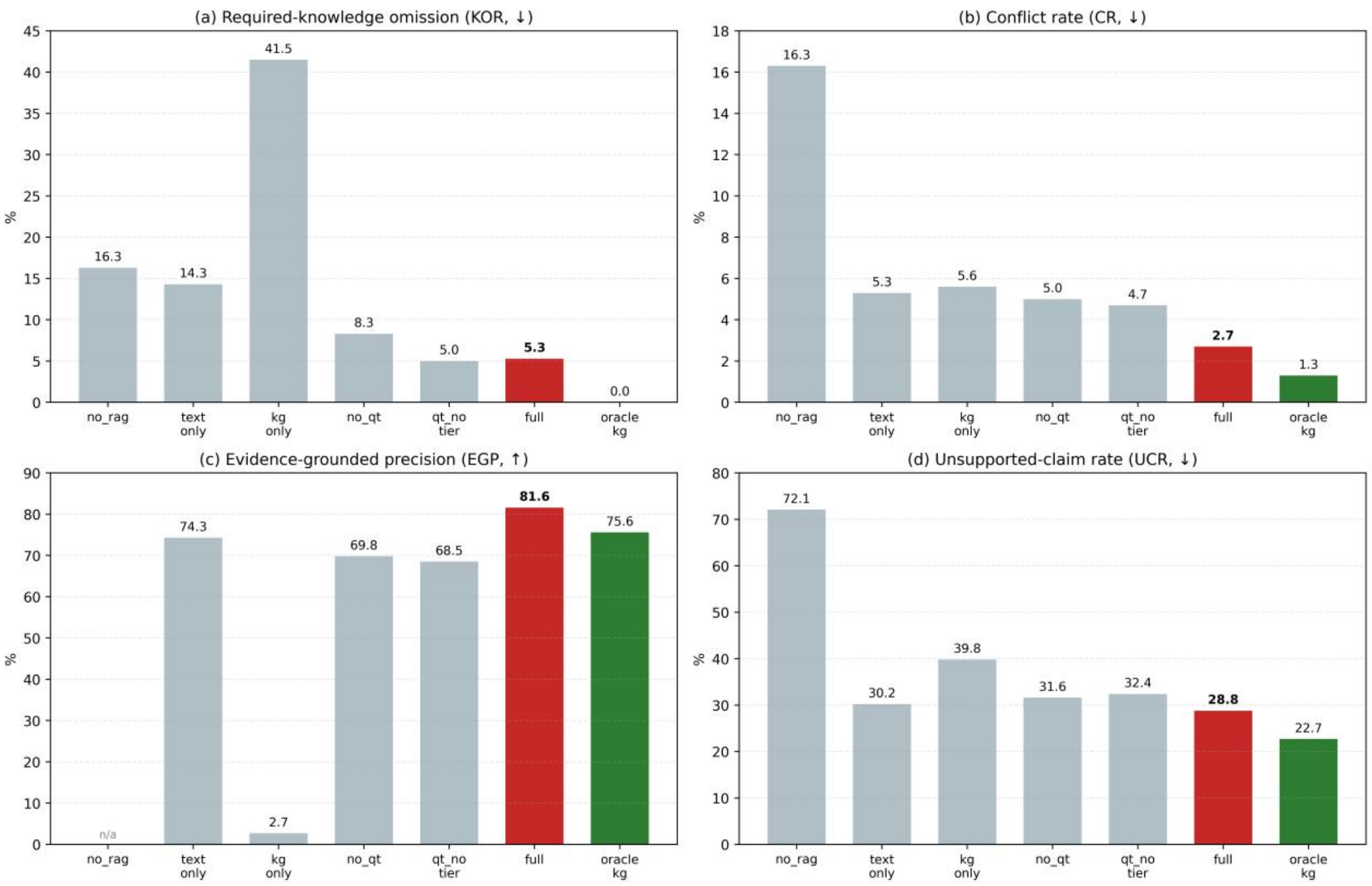

Figure 7. Guideline-based QA evidence quality across modes. (a) Required-knowledge omission rate (KOR, ↓) and (b) conflict rate (CR, ↓) on the 301 KG-anchored questions; (c) evidence-grounded precision (EGP, ↑); and (d) unsupported-claim rate (UCR, ↓) on the full 440-question set. Compared with 'no_rag', the 'full' framework reduces lower-is-better metrics and improves EGP; it also moves toward the oracle-evidence upper bound.

### *5.4 Ablation of quality-weighted retrieval and tiered prompting*

The two quality-propagation interfaces contribute through different channels. Adding quality-weighted retrieval ('no_qt' → 'qt_no_tier') lowers omission from 8.3% to 5.0% (McNemar $p = 0.031$), indicating that $Q(t)$-weighted ranking improves the downstream coverage of required evidence. Adding tier-conditioned prompting ('qt_no_tier' → 'full') further reduces the conflict rate numerically from 4.7% to 2.7%; taken alone, this incremental step is directional but not statistically significant given the small number of discordant pairs (McNemar exact $p = 0.180$, n = 301), whereas the cumulative reduction relative to 'no_rag' (16.3% to 2.7%) is significant ($p < 0.001$; Section 5.3). The two effects are therefore best interpreted jointly rather than as two independently significant steps: the full framework preserves low omission while achieving the lowest observed conflict rate, supporting the downstream reuse of construction-time quality for evidence selection while suggesting an additional benefit from reliability-aware evidence presentation.

### *5.5 Clinician evaluation*

Blinded ratings by two clinicians show a consistent ordering of overall answer quality: ‘no_rag’ 4.21 < ‘full’ 4.68 < ‘oracle_kg’ 4.80. The ‘full’ framework significantly outperforms the no-retrieval baseline ($p$ < 0.001, Wilcoxon), as does ‘oracle_kg’ ($p$ <0.001). For ‘full’ versus ‘no_rag’, significant improvements are also observed in accuracy, completeness, and hallucination control. Inter-rater ratings show moderate positive correlations (Pearson $r$ = 0.53-0.65), with 86-97% of paired ratings differing by no more than one point. The clinician assessment is consistent with the automatic omission- and conflict-rate trends, providing an independent cross-check of the automatic indicators.

### *5.6 Cross-model robustness*

To test whether the improvements are specific to the primary generator, we re-ran the 60-question cross-LLM subset with two additional generators (Qwen-Max and DeepSeek-V3) under identical retrieval, prompting, and judging settings. Both metrics show the same directional improvement across all three generators (Figure 8). Required-knowledge omission decreases from no-retrieval to full for every generator — GLM-4-Plus 18.3% → 6.7%, Qwen-Max 28.3% → 10.0%, and DeepSeek-V3 16.7% → 6.7%. Notably, the generator with the highest no-retrieval omission rate (Qwen-Max) shows the largest reduction, while all three generators reach a comparable low omission range (6.7-10.0%). Conflict rate likewise decreases for every generator — GLM-4-Plus 25.0% → 3.3%, Qwen-Max 15.0% → 10.0%, and DeepSeek-V3 20.0% → 5.0%. The GLM-4-Plus subset results follow the same direction as the full-set results (16.3% → 5.3% omission and 16.3% → 2.7% conflict). These results suggest that the benefits of construction-to-inference quality propagation are not specific to the primary generator.

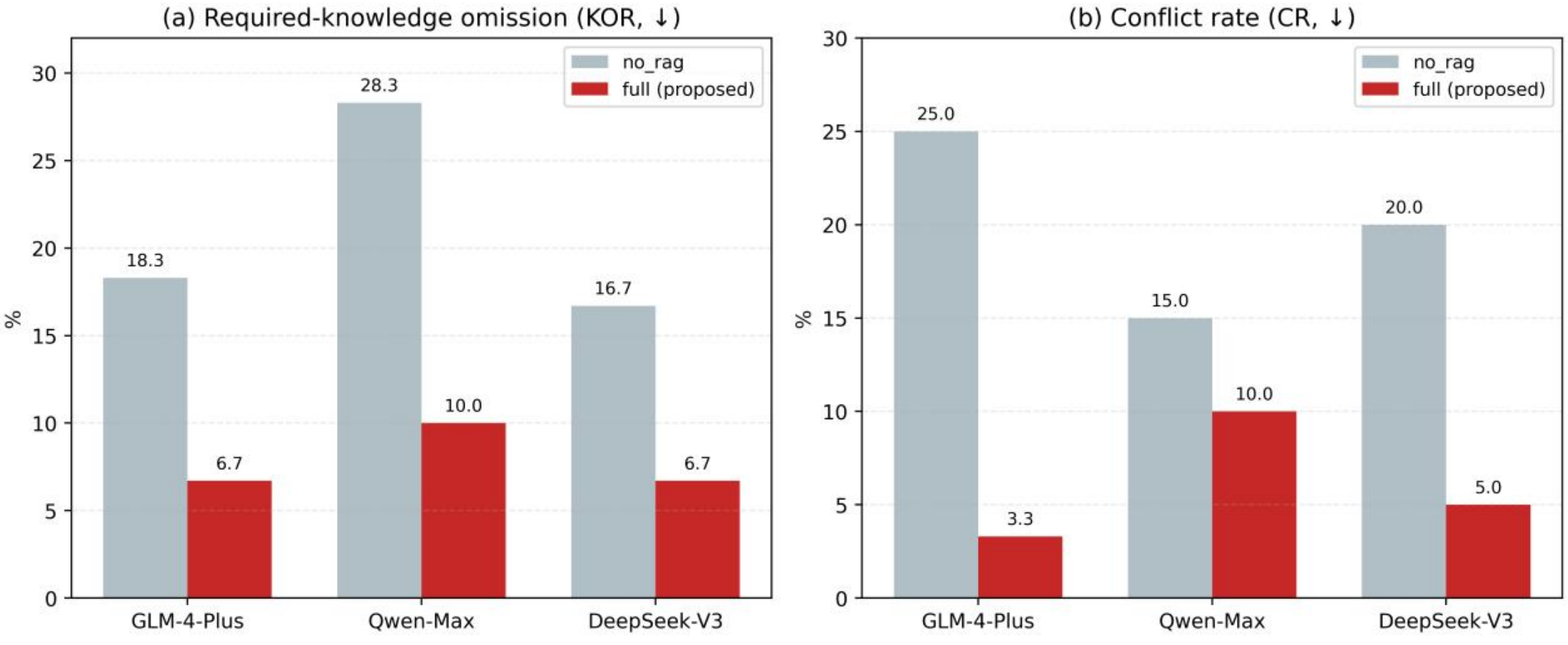

Figure 8. Cross-LLM evaluation on the 60-question subset. The ‘full’ framework reduces (a) required-knowledge omission and (b) conflict rate relative to ‘no_rag’ across GLM-4-Plus, Qwen-Max, and DeepSeek-V3, showing the same directional improvement across all three generators.

*5.7 Case study and provenance analysis*

We illustrate the end-to-end quality-propagation mechanism with a representative evaluation question (Q262): "*Which drugs should patients with hyperuricemia use with caution*?"

Entity linking identifies "patients with hyperuricemia" (高尿酸血症患者) as the seed entity (exact match, score = 1.0) and expands a two-hop candidate subgraph. Quality-weighted retrieval returns two core-tier triples ranked by Eq. (2): $t_1$ = (patients with hyperuricemia, *not suitable for*, niacin-type lipid-lowering drugs) ($Q(t)$ = 0.90, *SchemaConf* = 1.00, TextID = t1178, two-model support) and $t_2$ = (patients with hyperuricemia, *use with caution*, diuretics) ($Q(t)$ = 0.90, *SchemaConf* = 1.00, TextID = t216, t852, two-model support). Dense passage retrieval additionally returns passage t852, which provides contextual qualifiers distinguishing cautionary use for hyperuricemia patients from contraindication for gout patients — a distinction not prioritized by entity-linked KG retrieval alone.

The tier-conditioned prompt presents $t_1$ and $t_2$ as core evidence. The generated answer correctly identifies both drug classes and explicitly cites guideline passage [T852]. Beyond the generator's explicit citation, the system-level provenance links retained in the graph provide additional traceability: $t_1 \rightarrow$ t1178 and $t_2 \rightarrow$ {t216, t852}, enabling a clinician to verify each asserted fact against its source guideline passage even when the generator does not cite all relevant TextIDs in the answer text. Figure 9 summarizes the complete five-step pipeline. This example illustrates how construction-time quality guides retrieval and tier-conditioned evidence use, while the retained triple-TextID links provide traceability from graph evidence to source guideline passages regardless of whether the generator produces explicit citations.

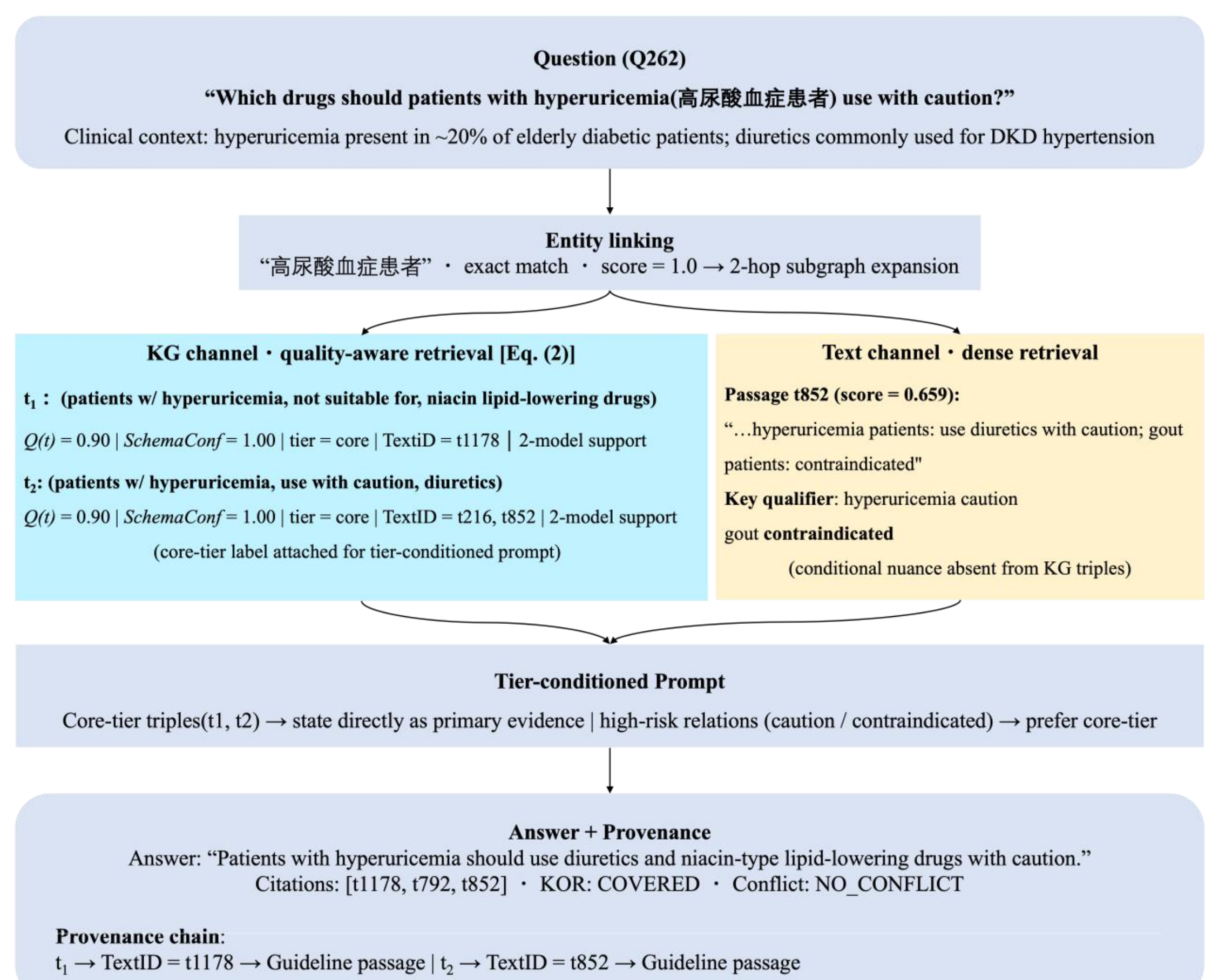

Figure 9. End-to-end quality propagation and provenance tracing for Q262 (hyperuricemia drug caution). Entity linking identifies the seed entity, after which quality-weighted KG retrieval provides two core-tier triples and dense passage retrieval supplies complementary guideline context, including the distinction between hyperuricemia caution and gout contraindication. The tier-conditioned prompt organizes evidence according to reliability for answer generation. The generator explicitly cites [T852]; system-level provenance additionally links each KG triple to its source passages ($t_1$ → t1178 and $t_2$ → {t216, t852}).

## *5.8 Error type analysis*

To examine how error composition varies with the quality-scoring signals, we analyze the 859 incorrect triples in the cross-version test set (1,888 labeled triples in total; Section 4.1). Each incorrect triple is assigned to one of five error categories through domain-expert review: (i) factual errors — the triple contradicts or lacks textual support in the source passage; (ii) entity errors — incorrect entity boundaries, entity confusion, or granularity mismatch; (iii) relation imprecision—the relation type deviates from the intended semantics without constituting a complete misuse; (iv) conditional generalization — knowledge applicable only under specific conditions, such as disease stage, patient subgroup, or treatment context, is extracted as an unconditional fact; and (v) direction errors — the head and tail entity roles are reversed.

Table 2. Error type distribution by *SchemaConf* quality class on the cross-version test set.

| Error type | *n* | Overall | Validated (precision=47.6%) | Candidate (precision=24.0%) | Uncertain (precision=0.6%) |
|---|---|---|---|---|---|
| Factual error | 500 | 58.2% | 58.0% | 59.3% | 50.0% |
| Entity error | 181 | 21.1% | 21.2% | 20.5% | 25.0% |
| Relation imprecision | 93 | 10.8% | 11.4% | 9.3% | 16.7% |
| Conditional generalization | 57 | 6.6% | 6.8% | 6.3% | 8.3% |
| Direction error | 28 | 3.3% | 2.6% | 4.6% | 0.0% |

Note: Column headers report the overall precision of each *SchemaConf* class. Percentages in the last three columns represent the distribution of error types among incorrect triples within each class (Validated: n=533; Candidate: n=302; Uncertain: n=24). The uncertain class is small, and its percentages should therefore be interpreted with caution.

As shown in Table 2, factual errors dominate the error set (58.2%), followed by entity errors (21.1%). The distribution of error types among incorrect triples is broadly similar across the three *SchemaConf* quality classes, despite substantial differences in overall precision. This pattern suggests that *SchemaConf* primarily stratifies the likelihood that a triple is correct, rather than selectively distinguishing among specific error types once a triple is incorrect. Direction errors show a somewhat different pattern, being more concentrated in the candidate class (4.6%) than in the validated class (2.6%), consistent with the intended role of the $s_{direction}$ component in penalizing structurally reversed triples.

A complementary view comes from examining error composition by model-support level, which determines *EvidScore* on the cross-version test set because $E_{human}$ and $E_{freq}$ are inactive (Section 3.4). Unlike the broadly stable pattern across *SchemaConf* classes, model support shows a clearer shift in error composition. Among low-support errors (support=1, n=565), factual errors dominate (65.0%), whereas relation imprecision accounts for only 7.3%. Among high-support errors (support=3, n=82), factual errors decrease to 40.2%, while relation imprecision (26.8%) and entity errors (31.7%) become more prominent; no conditional-generalization errors are observed. This pattern suggests that errors shared across multiple extraction models are more often associated with relation selection or entity representation, whereas low-support errors are more frequently factual in nature. Error composition therefore varies with model support in a manner different from that observed across *SchemaConf* classes, further supporting the complementary roles of structural and evidential quality signals. Given the small number of high-support errors (n=82), however, this pattern should be interpreted cautiously.

Two implications follow. First, the dominant factual-error category cannot be reliably detected by schema-conformance checking alone, because a factually incorrect triple may still conform to entity, relation, type, and direction constraints. Reducing such errors therefore requires evidential verification or other forms of factual validation in addition to structural scoring. Second, conditional generalization, although less frequent (6.6%), may carry substantial clinical risk because context-specific knowledge can be transformed into apparently universal statements. Passage-level provenance in the current framework allows such triples to be checked against their original context, but explicit representation of contextual conditions within the knowledge representation remains an important direction for future work.

### *5.9 Synthesis: conditions for effective quality propagation*

Across Sections 5.1-5.8, the results show that construction-time quality propagation operates through two complementary interfaces: quality-weighted retrieval improves required-evidence selection, while tier-conditioned prompting provides a mechanism for presenting evidence according to reliability. The overall gains are also observed across different answer generators, suggesting that the benefits of quality propagation are not restricted to a single model.

The effectiveness of the quality signal depends on three conditions being jointly satisfied: (i) coverage — the required evidence exists in the retained graph; (ii) activation — the retrieval pipeline successfully exposes and retains the relevant evidence in the final retrieved context; and (iii) reliability — the associated quality information, $Q(t)$ and $\tau(t)$, is sufficiently informative to guide evidence ranking and presentation. In this evaluation, the coverage condition is satisfied by design: all KG-anchored questions were constructed from existing evidence paths in the graph (Section 4.2), so all triples in the predefined required evidence path are present in the KG for every question in this set. Within this setting, the remaining KOR failures therefore arise from the activation/retrieval or generation stages rather than from graph coverage. We traced all 16 KOR failures in the full mode: 15 of 16 (93.8%) are activation/retrieval failures, in which entity linking, subgraph expansion, or subsequent ranking failed to surface the target triple in the top-15 retrieved KG evidence, typically because the head entity of the required path was not directly matched by a question term (e.g., multi-hop chains in which an intermediate node was missed, or long-tail triples whose seed entities were dominated by high-degree neighbors during subgraph expansion). The remaining failure (6.2%) is a generation failure, in which the required triple appeared in the retrieved context but was omitted from the generated answer. The concentration of failures at the activation/retrieval stage indicates that improving entity linking recall and subgraph expansion or ranking strategies is the most direct target for further gains.

## 6. Discussion

### *6.1 Construction-time quality as a persistent inference signal*

Many knowledge-grounded QA pipelines treat a constructed knowledge graph as a static store of facts, leaving retrieval relevance as the primary inference-time criterion. Our results support a different view: retrieval relevance and knowledge reliability are distinct dimensions, and construction-time quality estimates contain information that query relevance alone cannot recover. Relevance measures how well a triple matches the question, whereas reliability reflects schema conformance, multi-model agreement, passage-level support frequency, and human verification. These properties are determined independently of future queries. A highly relevant triple may still be structurally inconsistent, insufficiently supported, or supported by only a single extraction model; $Q(t)$ preserves these quality characteristics for later evidence selection. As shown in Sections 5.4 and 5.9, this additional reliability information is actionable: quality-weighted retrieval improves evidence selection, while tier-conditioned prompting provides a mechanism for presenting and using evidence according to reliability.

The practical value of this design is its lightweight integration. $Q(t)$ is a scalar attribute computed during construction and stored with the graph. Inference requires only quality-aware ranking and tier-conditioned prompting, without joint training, shared parameters, or query-time quality re-estimation. This enables integration into existing clinical-guideline QA pipelines, and the consistent results across generators (Section 5.6) indicate that the mechanism is not dependent on a specific downstream model.

Importantly, $Q(t)$ is not intended as a universal measure of triple quality or a guarantee of answer correctness. Rather, it provides an actionable construction-time signal for evidence selection and presentation. Under our evaluation setting, preserving and reusing this signal provides measurable benefits for downstream evidence selection and guideline-based QA, while remaining complementary to query-dependent relevance.

### *6.2 Conditions limiting quality propagation gains*

The analysis of the remaining QA failures in Section 5.9 reveals a practically important asymmetry: within the KG-anchored evaluation, the remaining required-knowledge omission failures are associated primarily with activation limitations rather than with graph coverage. This distinction carries a concrete engineering implication: adjusting quality estimation or tier thresholds addresses the reliability dimension, but cannot compensate for cases where relevant evidence is not surfaced through entity linking and subgraph retrieval. Indeed, 15 of the 16 remaining KOR failures in the full mode occur at the activation/retrieval stage, indicating that further gains depend primarily on improving evidence exposure rather than quality scoring alone.

It is important to note, however, that activation dominates in this evaluation because the question set was constructed from existing KG evidence paths, so the coverage condition is guaranteed by design. In broader deployment settings where questions are not constrained by known KG paths, coverage gaps are likely to become more prominent because required knowledge may be absent from the graph, and no retrieval or quality-ranking strategy can recover evidence that does not exist. Extending the framework to broader use therefore requires attention to KG coverage through additional guideline sources, broader knowledge extraction and expert review, and continual graph maintenance, alongside improvements in entity linking and subgraph retrieval.

The role of the schema component (*SchemaConf*) also requires a scoped interpretation. On the in-distribution calibration set, the ranking produced by $Q(t)$ remains close to that produced by *EvidScore* alone, with *SchemaConf* providing limited additional discrimination (Section 5.2). Its value becomes more apparent under distribution shift—on the cross-version, cross-model test set, where $Q(t)$ (AUC 0.748) outperforms both *EvidScore* (0.703) and *SchemaConf* (0.645) alone. We therefore view *SchemaConf* as a complementary structural constraint whose added discriminative value becomes more apparent under guideline revisions and extraction-model changes, rather than as a universal source of gains in stationary settings. More broadly, quality propagation does not replace knowledge acquisition or retrieval; it improves the use of reliable evidence when the upstream pipeline provides sufficient coverage and activation.

### *6.3 Implications for clinical knowledge systems and deployment*

For guideline-facing clinical assistants, practical utility depends not only on answer accuracy but also on evidence-grounded traceability: supporting evidence for asserted facts should be traceable to specific guideline passages, and uncertain evidence should be explicitly marked. Our framework operationalizes this principle through triple → TextID → passage provenance and reliability-tiered evidence presentation. In our evaluation, the full framework reduces unsupported claims and conflicts while preserving source-level traceability (Section 5.3). In a clinical decision-support context, such traceability may be important for clinical adoption because clinicians can verify the underlying guideline evidence and lower-reliability evidence can be presented as supplementary rather than definitive. The tiering mechanism is particularly relevant for high-risk relations, for which lower-tier evidence can be assigned a reference-only role rather than presented as definitive support (Section 3.8).

Deployment is also lightweight: the quality signal is computed during construction and stored with the graph, while inference requires only quality-aware ranking and prompt organization. The framework can support incremental maintenance by allowing affected triples to be re-scored when guideline sources are revised, rather than requiring complete graph reconstruction. Key practical challenges therefore include upstream knowledge coverage and reliable evidence activation through entity linking and graph retrieval, rather than solely increasing inference-time complexity, highlighting these components as priorities for future clinical deployment.

### *6.4 Limitations and future work*

Several limitations bound the scope of our conclusions. First, the study is single-domain and single-language: the graph, ontology, and QA set all target Chinese diabetes clinical guidelines. Within this scope, the cross-version, cross-model construction evaluation (Section 5.2) and the cross-generator QA evaluation (Section 5.6) provide partial evidence of robustness to guideline updates, extraction-model changes, and generator variation. However, the reported results should not be interpreted as evidence of language- or disease-level generalization, and the construction-to-inference design, although intended to be transferable, requires replication in other clinical domains to establish broader applicability. Second, generation uses a single primary generator (GLM-4-Plus); we probe generator dependence with a 60-question cross-LLM subset using Qwen-Max and DeepSeek-V3 (Section 5.6), where similar trends are observed, but a full multi-generator evaluation over the entire QA dataset remains future work. Third, the QA evaluation set is self-constructed, which raises external-validity concerns; we partially address this concern through blinded dual-clinician evaluation, denominator-anchored metrics whose rates are computed over predefined evidence and answer-claim populations rather than selected cases, and an oracle-evidence condition that provides an upper-bound reference for performance when the predefined required KG evidence is directly available. We cannot fully exclude a degree of overlap between question construction and knowledge-graph construction, since questions were built from existing KG evidence paths and refined with clinician input; this may favor evidence coverage relative to an independently sourced question set.

Fourth, the schema component's advantage is demonstrated on a single distribution-shift instance (one cross-version, cross-model test set); establishing broader robustness requires additional shift conditions. Fifth, in the main evaluation GLM-4-Plus serves simultaneously as one of the three extraction models, the primary answer generator, and the required-knowledge omission judge; although the cross-LLM subset (Section 5.6) holds the omission judge fixed while varying the generator, we cannot fully rule out self-consistency bias in the primary results. Sixth, the correctness labels underlying the calibration set, the cross-version test set, and the human-reviewed construction corpus were produced through iterative domain-expert review without a separately reported inter-annotator agreement statistic; formalizing this protocol (e.g., reporting Cohen's or Fleiss' $\kappa$ across annotators) would strengthen the evidentiary basis of the gold labels.

Finally, our evidence linking construction-time quality to answer quality is indirect, relying on retrieval and prompting ablations rather than demonstrating a direct monotonic relationship between triple quality and answer quality. The available QA set did not provide sufficient variation in evidence quality to estimate such a relationship reliably. Future work includes multi-disease and multi-language replication, controlled multi-generator evaluation, broader distribution-shift testing for *SchemaConf*, continuous KG maintenance under guideline revisions, and larger prospective clinical studies to assess deployment effectiveness beyond the laboratory setting.

## 7. Conclusion

We presented a quality-aware framework that propagates construction-time triple quality into guideline-based medical question answering. Rather than treating knowledge-graph construction and QA as disconnected stages, the framework computes a fused quality score $Q(t)$ during construction, organizes retained triples into reliability tiers, and reuses the same quality information during inference for evidence retrieval and tier-conditioned prompting. This construction-to-inference propagation requires no joint training and can be integrated into existing guideline pipelines through quality attributes stored with the graph.

Experiments demonstrate that tiered retention preserves substantially more correct knowledge than schema-based hard filtering while retaining uncertainty information instead of discarding potentially useful boundary knowledge. The full framework reduces required-knowledge omission (16.3% → 5.3%) and conflicting outputs (16.3% → 2.7%), while achieving 81.6% evidence-grounded precision with near-zero invalid citations. Clinician evaluation further shows a consistent quality ordering (no-retrieval 4.21 < full 4.68 < oracle-evidence 4.80). Ablation results indicate that quality-weighted retrieval and tier-conditioned prompting contribute through complementary channels: the former improves evidence selection, while the latter provides a mechanism for presenting evidence according to reliability.

Our central, scoped claim is that construction-time quality can serve as an operational signal for improving evidence selection and presentation in downstream medical QA—not that higher quality scores alone guarantee better answers, nor that any single quality measure is universally optimal. Within a single clinical domain and language, and under a KG-anchored evaluation setting in which the required evidence is present in the runtime graph, the framework enables evidence-grounded and traceable QA while highlighting graph coverage and evidence activation through entity linking and graph retrieval as key priorities for further improvement.

## Data Availability

An anonymized review package containing de-identified construction and evaluation data, frozen evaluation labels, analysis scripts, machine-readable results, documentation, and integrity checksums is available through a private peer-review link: https://figshare.com/s/5d29215612e1a1ff7761. The package supports independent recomputation of the reported analyses without redistributing copyrighted clinical-guideline text or clinician free-text comments. The full extracted triple dataset, complete annotated QA evaluation data, and additional analysis and figure-generation code will be deposited in a public repository upon acceptance. The source clinical guidelines are publicly available from their original publishers, with the full source list provided in Supplementary Table S1. No patient data were collected or used in this study.

## Declaration of generative AI and AI-assisted technologies in the manuscript preparation process

During the preparation of this work, the authors used ChatGPT (OpenAI) for language editing and to improve the clarity of the manuscript. After using this tool, the authors reviewed and edited the content as needed and take full responsibility for the content of the published article.

## Funding

This work was supported by the National Key Research and Development Program of China (Grant No. 2023YFC3605800), the National Natural Science Foundation of China (Grant No. 82300918), the Joint Open Research Fund of Suzhou Institute of Nanotechnology and Nano-bionics & Jiangsu Province Hospital, and the Medical Engineering Translational Fund of Jiangsu Province Hospital (Grant No. NM202408). The funders had no role in the study design, data collection and analysis, interpretation of the results, preparation of the manuscript, or decision to submit the article for publication.